%% file: emnlp2022.tex
\pdfoutput=1

\documentclass[11pt]{article}

\usepackage{EMNLP2022}

\usepackage{times}
\usepackage{latexsym}
\usepackage{url}
\usepackage[super]{nth}
\usepackage[utf8]{inputenc}

\usepackage[T1]{fontenc}

\usepackage[utf8]{inputenc}
\usepackage{tablefootnote}

\DeclareFontFamily{U}{rcjhbltx}{}
\DeclareFontShape{U}{rcjhbltx}{m}{n}{<->rcjhbltx}{}
\DeclareSymbolFont{hebrewletters}{U}{rcjhbltx}{m}{n}
\DeclareMathSymbol{\betdagesh}{\mathord}{hebrewletters}{129}

\usepackage{microtype}

\usepackage{inconsolata}

\usepackage{graphics}
\usepackage{graphicx}

\usepackage{bm}
\usepackage{amsthm}
\usepackage{amsmath}
\usepackage{amssymb}
\usepackage{multirow}
\usepackage{amsfonts}
\usepackage{enumitem}
\usepackage[normalem]{ulem}
\usepackage{color, colortbl}
\definecolor{Gray}{gray}{0.95}

\usepackage{booktabs}
\usepackage[]{caption}
\DeclareCaptionLabelFormat{AppendixTables}{A.#2}

\usepackage{dsfont}
\usepackage{amsmath}

\DeclareMathOperator*{\mean}{mean}
\DeclareMathOperator*{\std}{stdev}

\definecolor{hookersgreen}{rgb}{0.0, 0.44, 0.0}
\definecolor{indiagreen}{rgb}{0.07, 0.53, 0.03}
\definecolor{islamicgreen}{rgb}{0.0, 0.56, 0.0}
\definecolor{kellygreen}{rgb}{0.3, 0.73, 0.09}
\definecolor{alizarin}{rgb}{0.82, 0.1, 0.26}

\usepackage[normalem]{ulem}
\usepackage{pifont}
\newcommand{\cmark}{{\color{kellygreen} \ding{51}}}
\newcommand{\xmark}{{\color{alizarin} \ding{55}}}

\usepackage{mathtools}
\usepackage[normalem]{ulem}
\useunder{\uline}{\ul}{}

\usepackage{bbm}
\usepackage{wrapfig}

\newcommand{\Cc}{\mathcal{C}}

\newcommand{\Oc}{\mathcal{O}}

\newcommand{\Vc}{\mathcal{V}}

\newcommand{\Xc}{\mathcal{X}}

\newcommand{\xv}{\mathbf{x}}
\newcommand{\yv}{\mathbf{y}}
\newcommand{\zv}{\mathbf{z}}

\newcommand{\thetav     }{\boldsymbol \theta     }

\newcommand{\scriptnumber}[1]{{\scriptsize (#1)}}

\usepackage{makecell}

\usepackage{xspace}
\newcommand{\modelname}{{{\textsc{RLPrompt}}}\xspace}

\title{\modelname: Optimizing Discrete Text Prompts\\
with Reinforcement Learning}

\author{
Mingkai Deng$^{1}$\thanks{~~Equal contribution. Code available at \url{https://github.com/mingkaid/rl-prompt}},~~
Jianyu Wang$^{2*}$,~~
Cheng-Ping Hsieh$^{2*}$,~~
Yihan Wang$^{2}$,~~
Han Guo$^{1}$,~~\\
{\bf Tianmin Shu$^{3}$,~~
Meng Song$^{2}$,~~
Eric P. Xing$^{1,4,5}$,~~
Zhiting Hu$^{2}$}\\
$^1$Carnegie Mellon University,~~ $^2$UC San Diego,\\ $^3$MIT,~~
$^4$Mohamed bin Zayed University of Artificial Intelligence,~~ $^5$Petuum Inc.\\
{\small 
{\tt \{mingkaid,hanguo\}@cs.cmu.edu, \{jiw102,c2hsieh,yiw007,zhh019\}@ucsd.edu}
}
}
\begin{document}

\maketitle

\begin{abstract}
Prompting has shown impressive success in enabling large pre-trained language models (LMs) to perform diverse NLP tasks, especially with only few downstream data.
Automatically finding the optimal prompt for each task, however, is challenging. Most existing work resorts to tuning \emph{soft} prompts (e.g., embeddings) which fall short of interpretability, reusability across LMs, and applicability when gradients are not accessible. \emph{Discrete} prompts, on the other hand, are difficult to optimize, and are often created by ``enumeration (e.g., paraphrasing)-then-selection'' heuristics that do not explore the prompt space systematically. This paper proposes \modelname, an efficient discrete prompt optimization approach with reinforcement learning (RL). \modelname formulates a parameter-efficient policy network that generates the optimized discrete prompt after training with reward. 
To harness the complex and stochastic reward signals from the large LM environment,
we incorporate effective reward stabilization that substantially enhances training efficiency. \modelname is flexibly applicable to different types of LMs, such as masked (e.g., BERT) and left-to-right models (e.g., GPTs), for both classification and generation tasks. Experiments on few-shot classification and unsupervised text style transfer show superior performance over a wide range of existing fine-tuning or prompting methods. Interestingly, the resulting optimized prompts are often ungrammatical gibberish text; and surprisingly, those gibberish prompts are transferrable between different LMs to retain significant performance, indicating that LM prompting may not follow human language patterns.
\end{abstract}


\section{Introduction}
\label{sec:intro}

\begin{table*}[t]
\vspace{-5pt}
\centering
{\renewcommand{\arraystretch}{0.9}
\setlength{\tabcolsep}{4pt}
\small
\begin{tabular}{@{}l c c c c c c c c@{}}
\toprule
{\bf Methods} & \makecell{{\bf Frozen}\\{\bf LMs}} & {\bf Automated} & \makecell{{\bf Gradient-}\\{\bf Free}} & \makecell{{\bf Guided}\\{\bf Optimize}} & \makecell{{\bf Few-}\\{\bf Shot}} & \makecell{{\bf Zero-}\\{\bf Shot}}  & \makecell{{\bf Transferrable}\\ {\bf b/w LMs}} & \makecell{{\bf Interpret-}\\{\bf -ability}}  \\ \midrule
Fine-Tuning & \xmark  & \cmark  & \xmark  & \cmark  & \xmark & \xmark & \xmark & \xmark \\
Manual Prompt  & \cmark  &  \xmark  &   \cmark  &  \xmark & \cmark & \cmark & \cmark & \cmark \\
Instructions & \cmark  &  \xmark  &   \cmark  &  \xmark & \cmark & \cmark & \cmark & \cmark \\
In-Context Demonstration & \cmark  & \cmark  & \cmark & \xmark  & \cmark &  \xmark & \cmark & \cmark \\
Soft Prompt Tuning & \cmark  & \cmark  & \xmark & \cmark &  \cmark & \xmark & \xmark & \xmark \\ 
Discrete Prompt Enumeration & \cmark  & \cmark  & \cmark & \xmark  & \cmark & \cmark &  \cmark & \cmark \\
AutoPrompt~\cite{shin2020autoprompt} & \cmark &  \cmark & \xmark  &  \cmark & \cmark & \xmark & \cmark & \cmark  \\  
\midrule
RLPrompt ({\bf Ours}) & \cmark  & \cmark  & \cmark & \cmark  & \cmark & \cmark &  \cmark & \cmark \\
\bottomrule
\end{tabular}
}
\caption{ \small
Comparison of different (prompting) paradigms for using pre-trained LMs on downstream tasks, in terms of several desirable properties. 
\emph{Gradient-Free} methods do not require gradient information from the prompted LMs, which may be inaccessible or expensive to compute.
\emph{Guided Optimize} means the optimization/search is guided by gradient or reward signals, which tends to be more efficient than otherwise (e.g., enumeration).
Prompts of discrete tokens (as opposed to embeddings) are often \emph{transferrable}/reusable by different LMs. 
Our approach with RL can optimize prompts using rewards without supervised data (\emph{zero-shot}).
\emph{Discrete Prompt Enumeration} selects the best prompt from a large number of candidates 
\cite[e.g., from paraphrasing or generation, ][]{jiang2020can, gao2021LMBFF, liu2021KATE, prasad2022grips}. 
\emph{AutoPrompt} \cite{shin2020autoprompt} uses gradients to edit the discrete prompt tokens. 
See \S\ref{sec:relatedwork} and Appendix \S\ref{appendix:related-work} for more discussion.
}
\label{tab:summary}
\end{table*}

Prompting has emerged as a promising approach to solving a wide range of NLP problems using large pre-trained language models (LMs), including left-to-right models such as GPTs \cite{radford2019GPT2, brown2020language} and masked LMs such as BERT \cite{devlin-etal-2019-bert}, RoBERTa \cite{liu2019roberta}, etc. Compared to conventional fine-tuning that expensively updates the massive LM parameters for each downstream task, prompting concatenates the inputs with an additional piece of text that steers the LM to produce the desired outputs. 
A key question with prompting is how to find the optimal prompts to improve the LM's performance on various tasks, often with only a few training examples.

One of the most popular 
scheme
is to
tune \emph{soft} prompts (i.e., continuous embedding vectors) as they are amenable to gradient descent \citep[][{etc.}]{li2021prefix, vu2021spot, gu2021ppt, liu2021ptuningv1, mokady2021clipcap, qian2022contrastiveprefix, an2022input}. However, 
the resulting prompts
are, by their nature, hard for humans to understand~\cite{khashabi2021prompt, lester2021promptuning,hambardzumyan2021warp} 
and incompatible for use with other LMs. Besides, the required LM internal gradients are often expensive to compute, or 
simply unavailable for LMs deployed with only inference APIs
(e.g., GPT-3). 
It is thus often desirable to 
use \emph{discrete} prompts which consist of concrete tokens from a vocabulary. However, their discrete nature renders the optimization very difficult. Previous work has typically relied on manual engineering 
\cite{petroni2019KB, brown2020language, schick2021exploiting, tam2021improving}, or selecting from multiple paraphrased/generated prompts
\cite{jiang2020can, gao2021LMBFF, liu2021KATE, prasad2022grips, hao2022bertnet}. 
AutoPrompt \cite{shin2020autoprompt} uses gradient information to edit the prompt tokens, which suffers from training instability as well as the same applicability issue as gradient-based soft prompting, showing limited effectiveness in practice.


This paper presents \modelname, a new discrete prompt optimization approach based on reinforcement learning (RL). This approach brings together a wide range of desirable properties for efficient use on diverse tasks and LMs (Table~\ref{tab:summary}). 
Crucially, rather than directly editing the discrete tokens, which has been difficult and inefficient, \modelname trains a policy network that generates the desired prompts. Discrete prompt optimization thus amounts to learning a small number of policy parameters which we set as an MLP layer inserted into a frozen compact model such as distilGPT-2 \cite{2019distilgpt2}. 
This formulation also allows us to employ off-the-shelf RL algorithms \citep[e.g., ][]{guo2021text} that learn the policy with arbitrary reward functions---defined either with available data (e.g., in few-shot classification) or other weak signals when no supervised data is accessible (e.g., in controllable text generation).

On the other hand, RL for prompt optimization poses new challenges to learning efficiency: the large black-box LM presents a highly complex environment that,
given the prompt (i.e., actions), goes through a long series of complex transitions (e.g., reading the input and inferring the output)
before computing the rewards. This makes the reward signals extremely unstable and hard to learn from. 
To overcome this difficulty, we propose two simple yet surprisingly effective ways to stabilize the rewards and improve the optimization efficiency.

Experiments on few-shot classification and unsupervised text style transfer show our approach improves over a wide range of fine-tuning and prompting methods (e.g., those described in Table~\ref{tab:summary}), and is robust to different modeling choices (e.g., verbalizers in classification).
The resulting discrete prompts also facilitate rich interpretations and analyses for new insights into LM prompting. 
In particular, the optimized prompts, though inducing strong task performance, tend to be gibberish text without clear human-understandable meaning, echoing recent research \citep{webson2021prompt, zhao2021calibrate, prasad2022grips} that LMs making use of prompts do not necessarily follow human language patterns. Perhaps surprisingly, those gibberish prompts learned with one LM can be used in other LMs for significant performance, indicating that those different pre-trained LMs have grasped shared structures for prompting.


\begin{figure*}
\vspace{-5pt}
    \centering
    \includegraphics[width=0.7\textwidth]{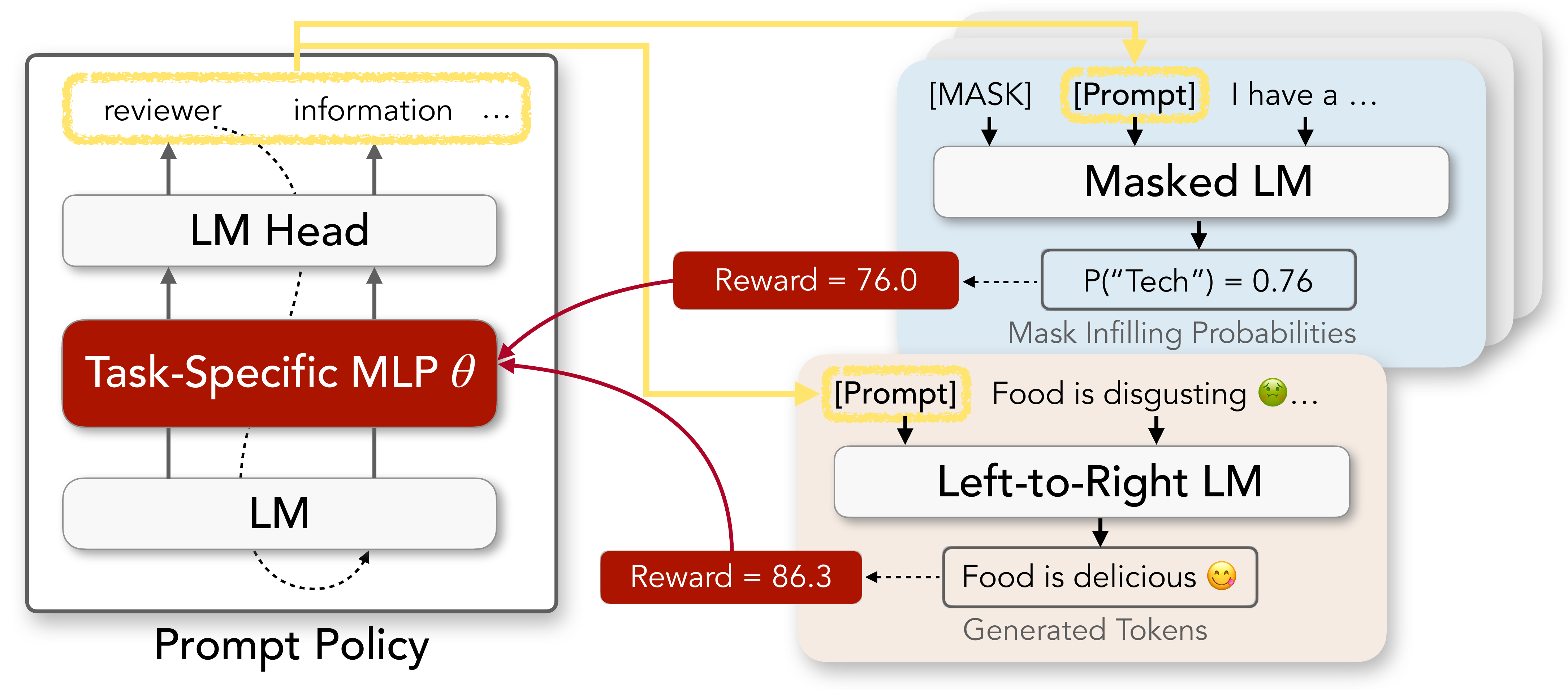}
    \caption{\small Overview of \modelname for discrete prompt optimization. All LMs (white boxes) are frozen. We build our policy network by training a task-specific MLP module inserted into a frozen pre-trained LM. 
    The figure above illustrates generation of a prompt (left), example usages in a masked LM for classification and a left-to-right LM for generation (top-right and bottom-right, respectively), and update of the MLP using RL reward signals. 
    }
    \label{fig:prompt-generator-implementation}
\end{figure*}

\section{Discrete Prompt Optimization with RL}
\label{sec:method}


We present \modelname, a framework for learning prompts of discrete tokens for pre-trained LMs to succeed in a wide range of NLP tasks.

As discussed in \S\ref{sec:intro}, discrete prompts can be easier to interpret and use
than continuous prompts, but also more challenging to learn due to intractable optimization over discrete tokens.
To solve this difficulty, we formulate discrete prompt optimization as an RL problem, using a continuous policy network to explore the prompt space. 
The network is highly parameter-efficient, only training a small MLP over a frozen compact LM (e.g., distilGPT-2).

Below, we present our RL formulation of discrete prompt optimization (\S\ref{subsec:method:formulation}-\ref{subsec:method:rl}).
We then discuss the design of our policy network (\S\ref{subsec:method:model}). 
Finally, we describe our reward engineering techniques to improve RL training (\S\ref{subsec:method:reward-engineering}). 


\subsection{Discrete Prompt Optimization Problem}
\label{subsec:method:formulation}

Extensive recent work \cite{brown2020language, jiang2020can, khashabi2021prompt, gao2021LMBFF} has shown it is possible to combine discrete text prompt $\zv$ with input $\xv$ to directly perform various NLP tasks using a pre-trained LM's generative distribution $P_{\text{LM}}(\yv | \zv, \xv)$, without needing to fine-tune the model.  
For instance, in classification, the LM can be a masked language model (MLM) such as BERT \cite{devlin-etal-2019-bert}, and $\yv$ is the class-label token (a.k.a. verbalizer like \texttt{positive} or \texttt{negative}) in the mask position; in a generation task, the LM can be a left-to-right model such as GPT-2 \cite{radford2019GPT2}, and $\yv$ is the generated text. See Figure~\ref{fig:prompt-generator-implementation} for illustrative examples. We use $\yv_\text{LM}(\zv , \xv)$ to denote the LM output on $\xv$ prompted by $\zv$.
%
%

Our goal is to find the optimal discrete prompt $\zv^*$ from vocabulary $\Vc$ to maximize some downstream performance measure $R$ of $\yv_\text{LM}(\zv^* , \xv)$.\footnote{Technically $\Vc$ can be any set of tokens. Here we simply use the downstream LM's vocabulary.}
The metric $R(\yv)$ can be as simple as match with gold label $\yv^*$ (e.g., in classification when data is available), but can also be more complex like the success criteria of controllable text generation, which composes aspects such as style accuracy, language quality, and content preservation. 
Assuming the prompts have fixed length $T$, we write the task of \emph{discrete prompt optimization} in the general format below: 
\begin{equation}
    \max\nolimits_{\zv \in \Vc^T} R\left(\yv_\text{LM}(\zv, \xv)\right).
\end{equation}
The optimization above, however, can be intractable because $\zv$'s discrete tokens are not amenable to gradient-based optimization, while brute-force search has the exponential complexity of $\Oc(|\Vc|^T)$. 
Previous work has to either approximate gradients over $\zv$ using continuous LM embeddings  \cite{shin2020autoprompt} or tweak human-written prompts with heuristics \cite{jiang2020can,mishra2021reframing,prasad2022grips}.

\subsection{The Reinforcement Learning Formulation}
\label{subsec:method:rl}
To overcome the difficulty, we formulate discrete text prompt optimization as an RL problem,
in which an agent selects prompt tokens $[z_1, \dots, z_T]$ one by one to maximize the reward $R(\yv_\text{LM}(\zv, \xv))$. At time step $t$, the agent receives previous prompt tokens $\zv_{<t}$ and generates the next prompt token $z_t$ 
according to a policy $\pi(z_t | \zv_{<t})$.
After the agent finishes the entire prompt $\hat{\zv}$, it receives the task reward $R(\yv_\text{LM}(\hat{\zv}, \xv))$.
Parameterizing the policy with $\thetav$, we can rewrite the problem above as
{\addtolength{\abovedisplayskip}{-7pt}
\addtolength{\abovedisplayshortskip}{-5pt}
\addtolength{\belowdisplayskip}{-7pt}
\addtolength{\belowdisplayshortskip}{-5pt}
\begin{equation}
    \max\nolimits_{\thetav} R(\yv_\text{LM}(\hat{\zv}, \xv)),\ \hat{\zv} \sim \prod\nolimits_{t=1}^T \pi_{\thetav}(z_t | \zv_{<t}).
\label{eq:rl}
\end{equation}
}
Compared to typical (soft) prompt tuning approaches, the RL formulation above has the key advantage of not needing gradient access to the LM,
treating it instead as a black-box function. 
This enables us to optimize prompts for LMs whose gradients are too expensive to compute, or LMs that are solely available as inference APIs (e.g., GPT-3). Compared to previous discrete prompt enumeration/paraphrasing, the RL approach explores the prompt space more efficiently guided by the reward signals.
%
The policy network also brings added flexibility, e.g., it can take other information such as the input $\xv$, leading to input-specific prompts (e.g., as used in text style transfer in \S\ref{subsec:method:reward-engineering}).  

During training, we explore the prompt space by sampling from the policy network. 
After the policy is trained, we select tokens greedily during inference to produce a deterministic prompt.
%
The reward objective in Eq.\eqref{eq:rl} can be optimized with any off-the-shelf RL algorithm. We use the latest soft Q-learning \citep[SQL,][]{guo2021text} which has shown advanced 
learning efficiency and performance on various text generation problems, with open-source implementation.\footnote{Our preliminary experiments indicate SQL often achieves superior performance than common policy gradient methods.}
Specifically, we use only its on-policy component. We refer interested readers to \citet{guo2021text} for more details.

\subsection{Efficient Parameterization of Policy}
\label{subsec:method:model}


We present an efficient parameterization of the policy network $\pi_{\thetav}$, which adapts a frozen pre-trained LM (i.e., policy LM) with a simple MLP layer that contains all the parameters $\thetav$ to be trained. The policy LM need not be the same as the LM we optimize the prompt for (i.e., task LM).
Figure~\ref{fig:prompt-generator-implementation} (left) illustrates the policy LM architecture.
Specifically, we use the LM to extract contextual embeddings of partial prompt $\hat{\zv}_{<t}$,
apply the added task-specific MLP layer to compute the adapted embeddings, and pass the output into the model's original LM head to obtain the next prompt token probabilities.
We describe more implementation details in Appendix \S\ref{appendix:implementation:policy}.
During training, we compute the MLP gradients by back-propagating through the policy LM. 
Our experiments (\S\ref{sec:experiments}) show that changing only the small set of MLP parameters is sufficient for producing strong performance.
After training, we discard the MLP and simply use the learned discrete text prompt for inference.


\subsection{Reward Engineering and Stabilization}
\label{subsec:method:reward-engineering}


Proper design of reward functions, a.k.a. reward engineering, is crucial to training efficiency and success in RL \cite{sutton2018reinforcement}. 
Discrete prompt optimization, in particular, poses new challenges due to its highly complex reward functions, which involve multiple steps (e.g., combining with input, passing through a black-box LM, and inferring the outputs),
each introducing its own variations. 
This makes the reward signal unstable and difficult to assess progress towards the task goal. 
To solve these difficulties, we propose two simple reward engineering techniques that effectively encourage and stabilize the RL training.

\paragraph{Input-Specific $z$-Score Reward}
\label{reward: z-score}
Different inputs can have different levels of difficulty for reasoning and prediction. Prompted LMs can thus see different reward scales 
for different inputs.
In text style transfer (\S\ref{subsec:tst-experiment}), for instance, some sentences may only require changing a few words to alter the style, so the LM naturally achieves higher rewards on them than on other more complex sentences.
Naively optimizing for all inputs with the same reward scale, therefore, can lead to training bias and instability. 
To mitigate this problem, we propose to use
input-specific $z$-score, which normalizes the rewards by input-specific means and standard deviations.  
This can be seen as a case of adaptive reward normalization, a commonly-used technique in RL~\cite{van2016learning}.
Formally, during prompt optimization, we sample a batch of prompts $Z(\xv)$ for each input $\xv$, and compute the reward $R(\yv_{\text{LM}}(\zv, \xv))$ for each prompt $\zv \in Z(\xv)$. 
After that, we compute the reward $z$-scores across prompts $Z(\xv)$. Using the shorthand $R_{\xv}(\zv)$ for $R(\yv_{\text{LM}}(\zv, \xv))$, namely the reward prompt $\zv$ receives for input $\xv$, we write the transformation as below:
\begin{equation}
    z\text{-score}(\zv, \xv) = \frac{R_{\xv}(\zv) - \mean\nolimits_{\zv' \in Z(\xv)} R_{\xv}(\zv')}{\std\nolimits_{\zv' \in Z(\xv)}R_{\xv}(\zv')}.
\end{equation}
To distinguish the $z$-scores of different inputs in the same batch, we condition our policy network on the inputs, i.e., $\pi_{\thetav}(\zv|\xv)$. 

\paragraph{Piecewise Reward}

If a reward function is misspecified or vulnerable, the policy may maximize it without moving towards the desired goal. For example, while learning classification using the ground-truth probability as reward function, the policy may find adversarial prompts~\cite{Wallace2019UniversalAdversarial,xu2022exploring} that lead to very high probabilities for a single class given arbitrary inputs. 
To overcome the issue, we propose to design piecewise reward functions~\cite{Yu2020MetaWorldAB, Rengarajan2022ReinforcementLW} with both smooth and disjoint components to better express the task priorities and improve robustness. 
Typically, we can include a dense, quantitative signal (e.g., label probability) to measure fine-grained progress towards the goal, and a sparse, qualitative signal only when certain states are achieved (e.g., certain accuracy on each class) by applying a large sudden increase in the reward. 
We illustrate an example design of piecewise reward in text classification (\S\ref{subsec:classifysetting}).

\section{Experiments}\label{sec:experiments}

The proposed \modelname is generally applicable to various types of LMs for performing different NLP tasks using diverse prompt formats (Figure~\ref{fig:prompt-generator-implementation}). We evaluate our approach on both classification (in few-shot setting, \S\ref{subsec:classifysetting}) and generation (unsupervised text style transfer, \S\ref{subsec:tst-experiment}), and perform rich analyses for new insights on LM prompting (\S\ref{subsec:analysis}).
We will release all code and data upon acceptance.

\subsection{Few-Shot Text Classification}
\label{subsec:classifysetting}

\input{tab-NLU-main-result}

Learning text classification with few labeled examples has been a problem of interest in many applications \cite{xu2018lifelong,yu2018diverse}.
We adopt the typical prompting setting~\cite{brown2020language,schick-schutze-2021-just} which solves classification by token infilling for an MLM like BERT or next-token prediction for a left-to-right LM like GPT-2. 
Classification, therefore, amounts to selecting tokens that correspond to a set of predetermined class labels, a.k.a., \emph{verbalizers} (e.g., ``\texttt{great}'' for positive sentiment and ``\texttt{terrible}'' for negative sentiment).
For instance, to classify the sentiment of an input sentence ``\texttt{food is delicious}'' using an MLM, we first fill our prompt and the input into a template ``\texttt{[Input] [Prompt] [MASK]}'', and then select the verbalizer token with the highest probability of filling into the \texttt{[MASK]} position.

\paragraph{Reward Function}
\label{reward:classification-piecewise}

The text classification task aims to correctly assign input text $\xv$ to its ground truth label $c$ from a set of classes $\Cc$. 
To mitigate the adversarial cases discussed in \S\ref{subsec:method:reward-engineering}, we design 
a piecewise reward function that 
encourages prompts to classify \textit{each} examples correctly. 
Given prompt $\zv$ and training example $(\xv, c)$, we compute the reward similarly to hinge loss as the gap between the label probability and the highest probability from other classes. Using the short hand $P_\zv(c) := P_{\text{LM}}(c|\zv, \xv)$ to denote the probability of label $c$, we can write the gap as $\text{Gap}_\zv(c) := P_\zv(c) - \max_{c'\neq c} P_\zv(c')$. 
The gap value is positive when the prediction is correct,
and negative otherwise. We denote $\text{Correct} := \mathbbm{1}[\text{Gap}_\zv(c) > 0]$. For a correct prediction, we multiply the positive reward by a large number to signal its desirability. 
The resulting reward function is as below:
\begin{equation}
    R(\xv, c) = \lambda_1^{1 - \text{Correct}} \lambda_2^{\text{Correct}} \text{Gap}_\zv(c),
\end{equation}
We describe more details and present ablations on reward design in Appendix~\S\ref{appendix:ablation}.

\paragraph{Datasets} Following previous work~\cite{gao2021LMBFF, sun2022black}, 
we experiment on a wide range of popular few-shot classification tasks including sentiment classification such as SST-2~\cite{socher2013recursive}, Yelp Polarity~\cite{zhang2015character}, MR~\cite{pang2005MR}, CR~\cite{hu2004CR}, SST-5~\cite{socher2013recursive}, and Yelp~\cite{zhang2015character}, and topic classification such as AG's News~\cite{zhang2015character}.
We additionally experiment on
Subj~\cite{pang2004sentimental},
TREC~\cite{voorhees2000building}, Yahoo~\cite{zhang2015character}, 
and DBPedia~\cite{lehmann2015dbpedia}, which we present in Appendix~\S\ref{appendix:implementation:fstc} due to space restriction.
We describe the dataset statistics in Table~\ref{tab:nlu-dataset} in the appendix. 
We train with $16$ examples per class, and validate using the same number of examples, in keeping with the standard few-shot setting~\cite{perez2021trueFS}.

\paragraph{Baselines}
We compare our approach with representative methods in the diverse training and prompting paradigms shown in Table~\ref{tab:summary}.
Additionally, we compare with the latest Black-Box (BB) Tuning~\citep{sun2022black}, which mixes discrete and soft prompts and tunes the soft part. We describe more details in Appendix \S\ref{appendix:implementation:fstc}.

\paragraph{Experiment Setup}
We use RoBERTa-large \cite{liu2019roberta} as our backbone model.
For our approach, we experiment with prompt lengths $T\in\{2,5\}$, and insert the prompt tokens at the same positions with our manual prompts \cite{schick2021exploiting, tam2021improving}.\footnote{It is known that increasing prompt length and/or inserting prompt tokens in multiple positions can often lead to improved performance. We leave further experiments to the future.} 
Please see Appendix \S\ref{appendix:implementation:classify} for more training details.
\paragraph{Results}
We present our few-shot classification results in Table~\ref{tab:cls-main}.
Our method (5 tokens) outperforms Manual Prompt and Instructions on all datasets, as well as In-Context Demonstration and Fine-Tuning on all but 1 and 2 datasets, respectively.
Compared to Prompt Tuning, our method achieves higher average accuracy with lower standard deviations, showing our approach is less sensitive to various training factors, a common issue for few-shot prompt tuning~\cite{li2021prefix, gu2021ppt}.
Our approach substantially outperforms BB Tuning with soft prompts, 
and is slightly better even after BB Tuning uses mixed discrete/soft prompts with 50 soft tokens.
Compared to previous discrete prompt optimization methods such as GrIPS~\cite{prasad2022grips} and AutoPrompt~\cite{shin2020autoprompt}, our method reaches superior accuracy on all benchmarks. 
On the additional datasets which tend to be multi-way (e.g., 16-class), Fine-Tuning shows higher performance, but our method continues the lead over prompting baselines, as we describe in more detail in Appendix~\S\ref{appendix:implementation:fstc}.

\paragraph{Training Efficiency}
To assess the training efficiency of our method, we compare our test accuracy across training steps with BB Tuning, which is also a gradient-free method but optimizes soft prompts. As Figure~\ref{fig:train-efficiency} shows, our RL-based method is as efficient as soft prompt tuning without access to LM gradients, converging in similar number of steps to BB Tuning, but with superior performance. Our training is also relatively stable, for even the worst prompts encountered after convergence perform comparably to BB Tuning on average.

\subsection{Unsupervised Text Style Transfer}
\label{subsec:tst-experiment}

\begin{table*}[t]
\vspace{-5pt}
\centering
{\renewcommand{\arraystretch}{1.0}
\small
\begin{tabular}{llllll}
\toprule
{Model}    & {Content}    & {Style}      & {Fluency}    & {\bf $\bm{J}$({\scriptsize C, S, F})} & {\bf GM({\scriptsize C, S, F})} \\
\midrule
\rowcolor{Gray}
\multicolumn{6}{l}{\textit{Training Baselines}}                                                                                                                                                             \\ 
Style Transformer & 75.2           & 96.4           & 58.6           & 46.1           & 75.2            \\
DiRR              & \textbf{78.8} & \textbf{97.7} & 75.6          & {59.6}          & {83.5}           \\
\rowcolor{Gray}
\multicolumn{6}{l}{\textit{Prompting Baselines (GPT-2-xl)}}                                                                                                                                             \\ 
Null Prompt       & 37.4          & 94.8           & 97.6           & 33.6           & 70.2            \\
Random Prompt     & 39.6           & 93.8           & \textbf{97.8} & 34.7           & 71.3           \\
Manual Prompt     & 64.2 \scriptnumber{6.8}          & 91.5 \scriptnumber{3.6}          & 93.2 \scriptnumber{1.4}          & 53.4 \scriptnumber{7.9}          & 81.8 \scriptnumber{3.4}           \\
\rowcolor{Gray}
\multicolumn{6}{l}{\textbf{\textit{\modelname (Ours)}}}                                                                                                                                                                    \\ 
distilGPT-2       & 57.3 \scriptnumber{1.7}          & 96.5 \scriptnumber{0.1}          & 85.3 \scriptnumber{1.3}          & 46.0 \scriptnumber{0.9}          & 77.9 \scriptnumber{0.4}           \\
GPT-2-small       & 60.0 \scriptnumber{0.4}          & 96.4 \scriptnumber{0.3}          & 89.0 \scriptnumber{2.8}          & 50.7 \scriptnumber{1.3}          & 80.1 \scriptnumber{0.8}           \\
GPT-2-medium      & 65.7 \scriptnumber{1.4}          & 95.2 \scriptnumber{1.2}          & 89.3 \scriptnumber{0.1}          & 56.1 \scriptnumber{1.0}          & 82.3 \scriptnumber{0.4}           \\
GPT-2-large       & 65.1 \scriptnumber{1.8}          & 94.6 \scriptnumber{2.3}          & 91.6 \scriptnumber{0.8}          & 56.5 \scriptnumber{1.3}          & 82.6 \scriptnumber{0.7}           \\
GPT-2-xl      & 72.1 \scriptnumber{1.5}          & 94.2 \scriptnumber{2.4}          & 89.5 \scriptnumber{0.5}          & \textbf{61.4 \scriptnumber{2.2}} & \textbf{84.7 \scriptnumber{1.0}}  \\
 \bottomrule
\end{tabular}
}
\caption{\small Automatic evaluation of our method vs. baselines on the Yelp \cite{shen2017style} sentiment transfer dataset. 
$J(\cdot)$ is our main metric which measures the average joint sentence-level scores of Content, Style, and Fluency as defined in \S\ref{subsec:tst-experiment}.
We also report the geometric mean (GM) of the three aspects.
Numbers in (parentheses) are standard deviations across 3 sets of prompts. 
}
\label{tab:tst-yelp-auto}
\end{table*}

Text style transfer (TST) \cite{jin2022deep} is a challenging problem, whose goal is to rewrite an input sentence into a desired style,
usually without supervised training data. For instance, in a sentiment transfer task,
given a negative sentence ``\texttt{The food is disgusting}'', the model should generate a positive sentence ``\texttt{The food is delicious}'', without training on such paired data.

Even without supervision data, our method can learn prompts with weak reward signals, which is not possible for most previous prompt optimization methods. 
Compared to previous TST work that trained models from scratch \citep[][{etc.}]{hu2017toward,shen2017style} or fine-tuned pre-trained LMs \cite{krishna-etal-2020-reformulating,liu2021DIRR,hu2021causal}, our method presents a more efficient solution that learns discrete prompts for a LM without updating the massive parameters.  


\input{fig-training-efficiency}

\paragraph{Reward Function}
Given input sentence $\xv$, the goal of TST is to generate output $\yv$ 
that preserves the information in $\xv$ while showing style attribute $s$. Following these priorities, we define the task reward as a simple sum of content preservation and target style intensity, described formally below:
\begin{equation}
    R(\xv, \yv, s) = \ \text{Content}(\xv, \yv) 
                       + \text{Style}(\yv, s).
\end{equation}
We implement the reward using common model-based metrics, described with more detail in Appendix \S\ref{appendix:implementation:tst}.
Because the reward shows different scales across inputs, we normalize the rewards using input-specific $z$-score as discussed in \S\ref{subsec:method:reward-engineering}, and present ablation studies on reward design along with our results.

\paragraph{Datasets}
Due to space restriction, in the main paper we evaluate on the popular Yelp sentiment transfer dataset \cite{shen2017style}. To further demonstrate our approach in few-shot setting, we include experiments on Shakespeare authorship transfer \cite{xu-etal-2012-shakespeare} in Appendix \S\ref{appendix:implementation:tst}.

\begin{table}[t]
\small
\centering
{\renewcommand{\arraystretch}{0.9}
\setlength{\tabcolsep}{3pt}
\vspace{6pt}
\begin{tabular}{lrrrr} \toprule
Model           & Content       & Style         & Fluency       & {\bf GM({\scriptsize C, S, F})} \\ \midrule
DiRR            & \textbf{4.83} & \textbf{4.69} & 4.64          & \textbf{4.72}                   \\
Manual Prompt   & 4.25          & 4.38          & \textbf{4.86} & 4.49                            \\
\modelname (Ours) & \underline{4.41}          & \underline{4.68}          & \underline{4.80}           & \underline{4.63}                            \\ \bottomrule
\end{tabular}
}
\caption{\small Human evaluation on Yelp on 5-Likert scale
where the best result on each aspect is {\bf bolded} and the second best result \underline{underscored}. DiRR relies on model fine-tuning.
}
\label{tab:tst-yelp-human}
\end{table}

\paragraph{Baselines}
We evaluate our method against both training and prompting baselines. We compare with two strong training methods, Style Transformer \cite{dai2019styleTransformer} and DiRR \cite{liu2021DIRR}. 
In particular, DiRR fine-tunes GPT-2 \cite{radford2019GPT2} with RL signals, which can be seen as a full-model tuning analogue to our method. 
For the prompting baselines, we compare with (1) Null Prompt, which does not use any prompt, (2) Random Prompt, which samples 5 tokens from the vocabulary as prompts, and (3) Manual Prompt, which averages the performance of 3 human-written templates, one by \citet{reif2021recipe} and two written for this experiment. 

\paragraph{Experiment Setup}
We experiment with GPT-2 of varying sizes, ranging from the smallest distilGPT-2 with 82M parameters to the largest GPT-2-xl with 1.5B parameters. We fix the prompt length $T=5$. 
To generate output $\hat{\yv}$, for all comparison methods, we sample 32
candidates from the respective models, and select the one with the highest reward.
More details are in Appendix \S\ref{appendix:implementation:tst}.

\begin{table*}[h]
\vspace{-5pt}
\setlength{\tabcolsep}{4pt}
\centering
{\renewcommand{\arraystretch}{0.9}
\small
\begin{tabular}{lrlllll}
\toprule
{Method} & {Prompt PPL}$\downarrow$                                         & {Content} & {Style} & {Fluency} & {$J$({\scriptsize C, S, F})} & {GM({\scriptsize C, S, F})} \\ \midrule
\modelname                 & $254$K {\scriptsize ($238$K)} &
\textbf{72.1 {\scriptsize(1.5)}}       & 94.2 {\scriptsize(2.4)}     & 89.5 {\scriptsize(0.5)}       & \textbf{61.4 {\scriptsize(2.2)}} & \textbf{84.7 {\scriptsize(1.0)}}  \\
+ Fluency                     & \textbf{82.1 {\scriptsize(2.4)}}                                                  & 52.4 {\scriptsize(1.5)}       & \textbf{96.2 {\scriptsize(0.9)}}     & \textbf{94.6 {\scriptsize(1.0)}}       & 46.7 {\scriptsize(0.7)} & 78.1 {\scriptsize(0.4)} \\ \bottomrule
\end{tabular}
}
\caption{ \small
Comparison of prompt optimization with fluency constraint vs no constraint on the Yelp dataset. Both experiments use GPT-2-xl as the text generation model. Prompt PPL is the prompt's perplexity under a GPT-2 langauge model. The text style transfer metrics are the same as in Table \ref{tab:tst-yelp-auto}.}
\label{tab:tst-fluency}
\end{table*}

\paragraph{Evaluation}
{\normalsize
Following previous work, we perform both automatic and human evaluation on the content preservation, style accuracy, and fluency of model outputs.
For automatic evaluation,
we measure \emph{Content} by the state-of-the-art input-output alignment~\cite{deng-etal-2021-compression} using pre-trained LM, \emph{Style} by fine-tuned style classifiers, and \emph{Fluency} by a grammaticality classifier~\cite{krishna-etal-2020-reformulating}.
To aggregate the quality dimensions, we average the joint sentence-level scores of examples $\xv$ from the test set $\Xc$,
strictly following
\citet{krishna-etal-2020-reformulating}'s 
protocol defined below:
}%
\vspace{-6pt}%
\begin{align}
    J(&\text{\small Content}, \text{\small Style}, \text{\small Fluency}) = \\ 
    &\text{mean}_{\xv \in \Xc} \left( \text{\small Content}(\xv) \cdot \text{\small Style}(\xv) \cdot \text{\small Fluency}(\xv) \right), \notag
\end{align}
which requires each sentence to preserve input content, have the correct style, and be fluent. 
We also report the geometric mean (GM) of the three overall aspect scores.
We conduct human evaluation for Yelp by rating 100 outputs from each model with 5 annotators.
We describe more evaluation metrics and results in Appendix \S\ref{appendix:implementation:tst}.


\begin{figure}[t]
    \centering
    \vspace{-2pt}
    \includegraphics[width=0.48\textwidth]{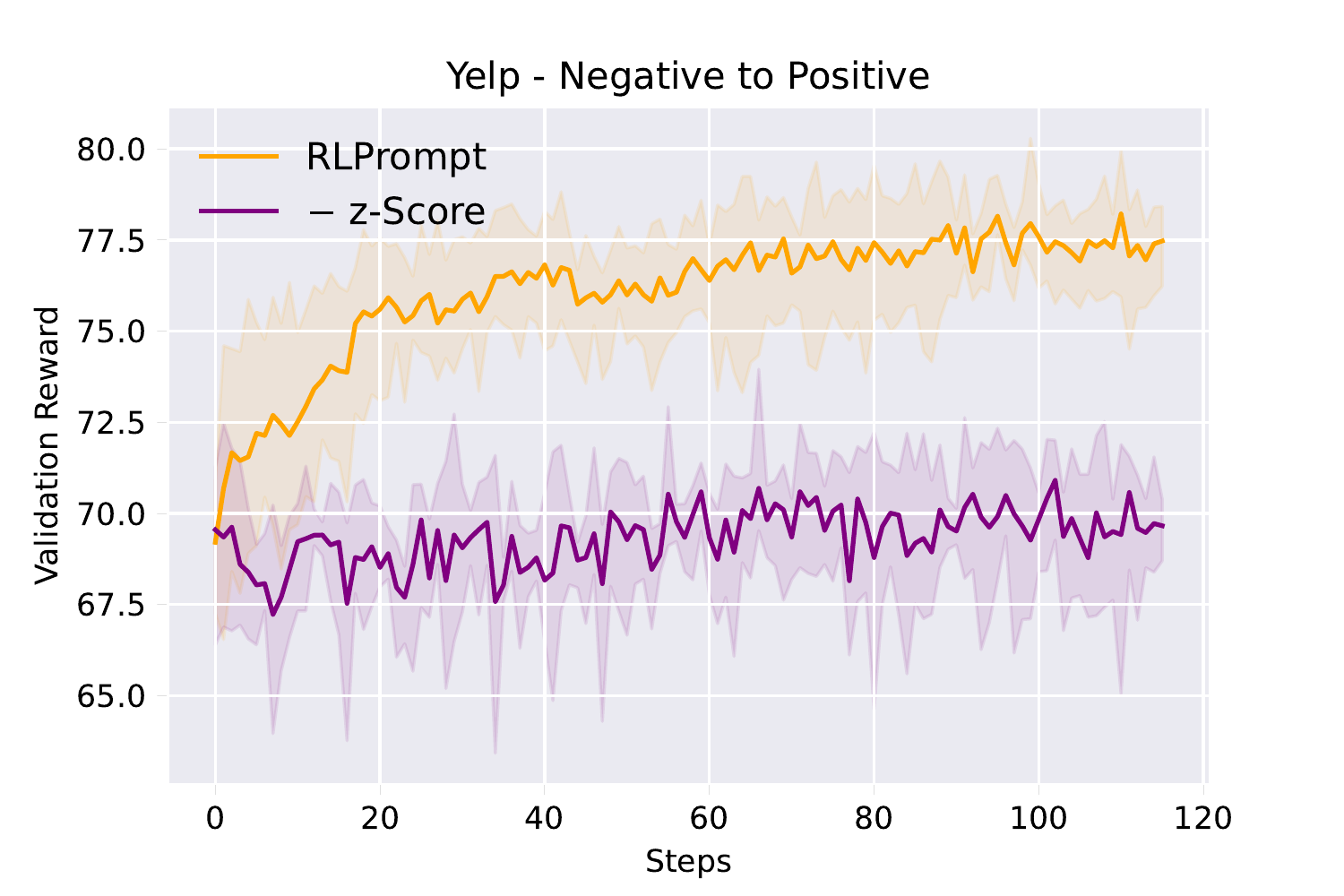}
    \vspace{-15pt}
    \caption{\small Comparison of our method with ({\color{orange} orange}) and without ({\color{purple} purple}) $z$-score reward normalization. The format is the same as Figure~\ref{fig:train-efficiency}. Additional comparisons are in Figure~\ref{fig:yelp-ablations-2}.}
    \label{fig:yelp-ablations}
\end{figure}

\paragraph{Results}
We present the automatic evaluation results for Yelp in Table \ref{tab:tst-yelp-auto}.
Compared to the expensive training baselines (Style Transformer and DiRR), our method with GPT-2-xl shows slightly lower content preservation and style accuracy, but have markedly better fluency, which leads to higher or competitive overall joint score $J(\cdot)$ and geometric mean GM$(\cdot)$. This may be because our method better preserves the LM's fluent generation capability by freezing its parameters.
Relative to prompting baselines, our optimization strongly improves the default performance.
In particular, our trained prompts performs better on average with lower variance than manual prompts, which sees performance vary wildly across prompts with similar meanings.
We present all manual and learned prompts along with their performance in Table \ref{tab:tst-prompt-examples} in appendix. 
Within our own method, we can see the performance increasing monotonically from the smallest distilGPT-2 to the largest GPT-2-xl. 
Human evaluation results (Table \ref{tab:tst-yelp-human}) show similar conclusions, where our method is competitive with the costly training method DiRR by obtaining slightly lower content and style scores but higher fluency. 
On Shakespeare, our method shows similar performance patterns even under the few-shot setting, which we discuss in more detail in Appendix \S\ref{appendix:implementation:tst}.


\begin{figure}[t]
    \vspace{-5pt}
    \centering
    \includegraphics[width=0.48\textwidth]{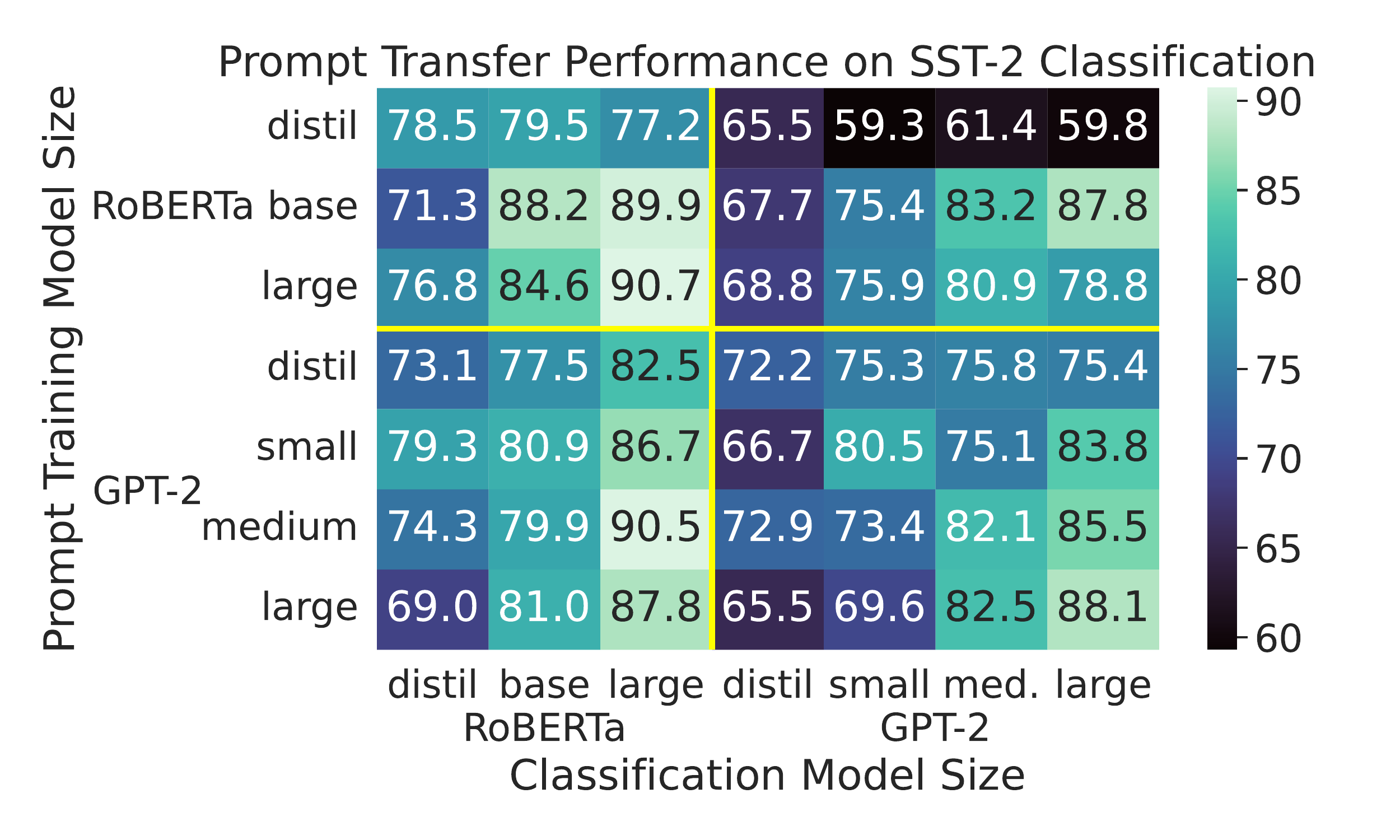}
    \\
    \vspace{-10pt}
    \caption{\small Heatmap of sentiment analysis performance with transferred discrete prompts of 2 tokens. The columns represent the models used to learn the prompts, and the rows represent the models we perform classification with. Brighter color represents higher accuracy.}
    \label{fig:prompt-transfer-sentiment-2tokens}
\end{figure}

\paragraph{Ablation Study}
As discussed earlier, we transform the reward function for TST using input-specific $z$-score to mitigate the training instabilities caused by the different scales of reward across inputs. 
To study the impact of this technique on RL training, we compare our training success with and without $z$-score normalization. Specifically, we test on Yelp \cite{shen2017style} using the distilGPT-2 model as an example. Following typical practice in RL, we run 5 experiments for each variant using different random seeds and compute the validation reward every 50 training steps. As the visualized results in Figures~\ref{fig:yelp-ablations} and \ref{fig:yelp-ablations-2} show, $z$-score normalization achieves both superior performance and more stable improvement across random seeds and training tasks. Because training easily collapsed without $z$-score using the original hyperparameters, we tuned the reward shaping scheme to transform a scale of [50,100] into [-50,50], which substantially improved training stability and results.

\subsection{Analysis}
\label{subsec:analysis}

\paragraph{Fluent vs. Gibberish Prompts}
We study the interaction of prompt fluency with downstream task performance, because fluent prompts are valuable for interpretability and insights into 
useful task instructions for LMs.
Our results show that \emph{good optimized prompts 
for the downstream task are often incoherent gibberish}.
For instance, one learned prompt for sentiment transfer is ``\texttt{Parameters Comparison
)=( Compare either}''.
The observation suggests that pre-trained LMs make use of prompts differently from humans, in line with previous discoveries in prompt-based fine-tuning~\cite{webson2021prompt}.
To understand how prompt fluency could impact the model performance, we 
evaluate on text style transfer (\S\ref{subsec:tst-experiment}). Specifically, we optimize \emph{fluent} prompts by constraining the prompt policy's action space (see Appendix~\S\ref{appendix:additional-analysis} for the constraint), and compare 
with our standard method (without fluency constraint) in Table \ref{tab:tst-fluency}. 
Results show that the fluency-constrained prompts have remarkably lower perplexity, which indicates higher language coherence. For instance, one fluent prompt we learned for to-positive transfer is ``\texttt{I love my life (}''.
However, these prompts receive much lower task performance in terms of $J(\cdot)$ and GM$(\cdot)$.
We present the learned fluent and gibberish prompts in Table \ref{tab:tst-prompt-examples} in the appendix.


\begin{table}[t]
    \centering
{\renewcommand{\arraystretch}{0.9}
\small
    \begin{tabular}{@{}lcc@{}}
    \toprule
    Verbalizers & \modelname & Manual \\ \midrule
    terrible, great & \textbf{92.8 \scriptnumber{0.8}} & 82.8 \\ 
    bad, good & \textbf{91.2 \scriptnumber{1.4}} & 79.7 \\
    negative, positive & \textbf{92.2 \scriptnumber{0.6}} & 76.8 \\
    \bottomrule
    \end{tabular}}
    \caption{\small Comparison of \modelname and manual prompt on SST-2 using different verbalizers. 
    }
    \label{tab:verbalizers}
\end{table}

\paragraph{Transferring Prompts across LMs}
\label{sec: Transfer across Models}
One unique advantage of discrete prompts over soft prompts is they are transferrable across models, due to the common text space instead of the model-specific latent space. 
This enables us to study the connections between different LMs by comparing the transfer performance of prompts trained from these models (e.g., taking a prompt trained on distilGPT-2, and applying it to GPT-2-xl). 
Interestingly, experiments show that the \emph{optimized prompts, though largely gibberish text, can indeed retain significant performance after transferring to different LMs}. Furthermore, \textit{prompts can transfer from smaller to larger models for similar or even better performance}.
More concretely, for this study, we use both few-shot classification (\S\ref{subsec:classifysetting}) and style transfer (\S\ref{subsec:tst-experiment}). Specifically for classification, we train prompts on various sizes of RoBERTa and GPT-2 and apply them to every other model for classification. We tabulate the average performance over 5 runs in the heatmap of Figure~\ref{fig:prompt-transfer-sentiment-2tokens}. 
Overall, all prompts can transfer between models, but the success depends on both the source and target LMs. 
For example, prompts learned from larger models see sharp performance declines when applied to smaller models, indicating that the structures they activate in large LMs may be less present in smaller ones. 
In contrast, prompts learned from smaller models reach similar or better performance on larger models (e.g., RoBERTa-base to -large).
Experiments on TST exhibit similar patterns as shown in Figure~\ref{fig:prompt-transfer} in Appendix~\S\ref{appendix:additional-analysis}. 
Perhaps surprisingly, prompts learned from MLMs like RoBERTa transfer well to left-to-right LMs like GPT-2 and vice versa, showing the LM structures they activate are largely shared across model types.
These findings open up a promising and exciting direction for future research---enabled by the transferrability across LMs, we may learn a prompt cheaply from smaller models, and apply it to a larger, more powerful model for inference.

\paragraph{Robustness to Classification Verbalizers}
It is known that prompted classification is sensitive to verbalizer choices. Manual design requires domain expertise and understanding of the base LMs. Previous research devised various methods for automatic verbalizer search~\cite{schick2020label, shin2020autoprompt, gao2021LMBFF}.
In few-shot classification, our method can discover well-performing prompts given a wide variety of verbalizers. Table~\ref{tab:verbalizers} shows the results on SST-2 with several intuitive verbalizers, 
averaged over 3 random seeds for each verbalizer pair. Across different verbalizers, our prompts consistently outperform manual prompt with smaller variation, showing our approach is robust to the choice of verbalizers. 
We report similar results on AG's News in Table~\ref{tab:verbalizers2} in the appendix.

\section{Related Work} 
\label{sec:relatedwork}

We discuss briefly the various prompting paradigms in previous work, and provide more comprehensive discussion in Appendix \S\ref{appendix:related-work}.
The conventional usage for pre-trained LMs is \textit{fine-tuning} on downstream datasets~\citep[\textit{etc.}]{devlin-etal-2019-bert, lewis2020bart}, which expensively updates all model parameters and shows limited success with small datasets. 
\citet{brown2020language} show that \textit{manual prompts} can steer large LMs to perform NLP tasks without any training \cite{raffel2020T5,schick2021exploiting,sanh2021T0}.
Another line of work~\cite{weller2020learning,efrat2020turking,mishra2021NI,wang2022NI2} develop \textit{instructional prompts} which provide task descriptions instead of fill-in-the-blank questions. 
With few-shot training examples, \citet{brown2020language} and follow-ups \cite{gao2021LMBFF,liu2021KATE,lu2021fantastically,min2022rethinking} achieve remarkable performance by inserting \textit{in-context demonstrations}. 
Replacing discrete prompts with continuous embeddings, several works \cite{qin-eisner-2021-learning,li2021prefix,liu2021ptuningv1} \textit{tune soft prompts} using gradient descent.
By their continuous nature, however, soft prompts are difficult to understand~\cite{lester2021promptuning,hambardzumyan2021warp,khashabi2021prompt}, require expensive gradient information \cite{sun2022black,diao2022black} and are incompatible for reuse across models due to mismatched latent spaces \cite{su2021transferability}. 
Some existing works seek to locate better discrete prompts by \textit{augmenting human-written prompts with heuristics} such as paraphrasing \cite{jiang2020can}, editing \cite{prasad2022grips}, and reframing \cite{mishra2021reframing}, and selecting by some downstream metric.
\textit{AutoPrompt}
\cite{shin2020autoprompt} edits discrete prompts with guidance from model gradients, which sees some success with large training data but limited general applicability due to unstable approximations.

\section{Conclusion}
We have presented \modelname, an efficient and flexible approach for discrete prompt optimization using RL, which 
improves over a wide range of fine-tuning and prompting methods in experiments on few-shot classification and unsupervised text style transfer. Analysis reveals that strong optimized prompts are incoherent but transferrable between LMs for remarkable performance. The observation opens up many promising possibilities for prompting, such as learning prompts cheaply from smaller models and performing inference with larger models. We are excited to explore further.

\section{Limitations}


While our prompt optimization method performs well on regular-sized LMs like RoBERTa and GPT-2, we have not experimented with more recent huge models like GPT-3 \cite{brown2020language}. As is the case for typical RL methods, designing reward functions may need domain expertise. However, we may solve this problem using techniques such as inverse RL, which learns the reward function from data. In terms of transferrability across models, we have not looked closely into the patterns of the learned prompts, or so-called ``secret language" of LMs. We look forward to studying all these questions in future work. 


\section*{Acknowledgements}
We thank all reviewers for their invaluable comments and feedback. 
Mingkai Deng and Han Guo are supported by US NGA NURI No. HM0476-20-1-0002 and the National Science Foundation under Grant No. IIS-15-63887, CCF-16-29559, IIS-16-17583, IIS-19-55532, CNS-20-08248, IIS-21-23952, and BCS-20-40381. Any opinions, findings, and conclusions or recommendations expressed in this material are those of the authors and do not necessarily reflect the views of the NGA or the U.S. Government.

\section*{Ethics Statement}
We acknowledge the ACL Code of Ethics and the ACM Code of Ethics and Professional Conduct and strictly adhere to the rules throughout the course of this research. We would like to note that 
massive pre-trained language models (with prompting or not)
could be used maliciously to generate fake, toxic, or offensive content. On the other hand, we hope the proposed prompting technique can be useful for harnessing and controlling the LMs from the unethical behaviors.


\bibliography{anthology,custom}
\bibliographystyle{acl_natbib}

\clearpage

\appendix

\section{Experiment Details}
\label{appendix:implementation}

\subsection{Policy Network}
\label{appendix:implementation:policy}
For all tasks, we uniformly use distilGPT-2 (\cite{2019distilgpt2}) with 82M parameters as a compact policy LM, and implement a generously parameterized MLP with 1 hidden layer and 2048 hidden states. Given distilGPT-2's hidden size of 768, we only add 3.1M parameters, or 3.8\% of the LM parameters. 

\subsection{Few-Shot Text Classification}
\label{appendix:implementation:fstc}

\paragraph{Reward Function Details}

During training, we compute the reward for prompt $\zv$ by averaging over all our few-shot training examples. We set the balancing weights $\lambda_1 = 180$ and $\lambda_2 = 200$ by tuning on the validation set.

\paragraph{Baseline Implementation Details}
For Manual Prompt, we take the hand-crafted prompts from \citet{schick2021exploiting}. 
For Instructions, we manually create task descriptions and label definitions following \citet{mishra2021NI}'s protocol (shown in Table~\ref{tab:nlu-instruction}) and prepend the instructions to the inputs.
For In-Context Demonstration~\cite{brown2020language}, we randomly select one training example per class and concatenate them with the input texts. 
For Prompt Tuning~\cite{lester2021promptuning}, 
we replace the Manual Prompt tokens with five soft tokens in the same positions for fair comparison, and optimize them using Adam optimizer with learning rate $1 \times 10^{-2}$ and batch size 16 for 400 epochs. 
For Black-Box Tuning~\cite{sun2022black} with mixed prompt, we use 50 soft tokens and 8,000 budget following the default setting. For its soft-prompt-only setting, we also optimize with the same budget.
For Fine-Tuning, we train with Adam optimizer with learning rate $1 \times 10^{-5}$ and batch size 16 for 100 epochs.
For Discrete Prompt Enumeration, we take GrIPS~\cite{prasad2022grips} as a state-of-the-art example. 
For AutoPrompt~\cite{shin2020autoprompt}, we use 5 prompt tokens and perform prompt search with a batch size of 16 using the few-shot training examples. 
For each baseline, we pick the model with the best validation accuracy for evaluation.

\begin{figure}[t]
    \centering
    \includegraphics[width=0.48\textwidth]{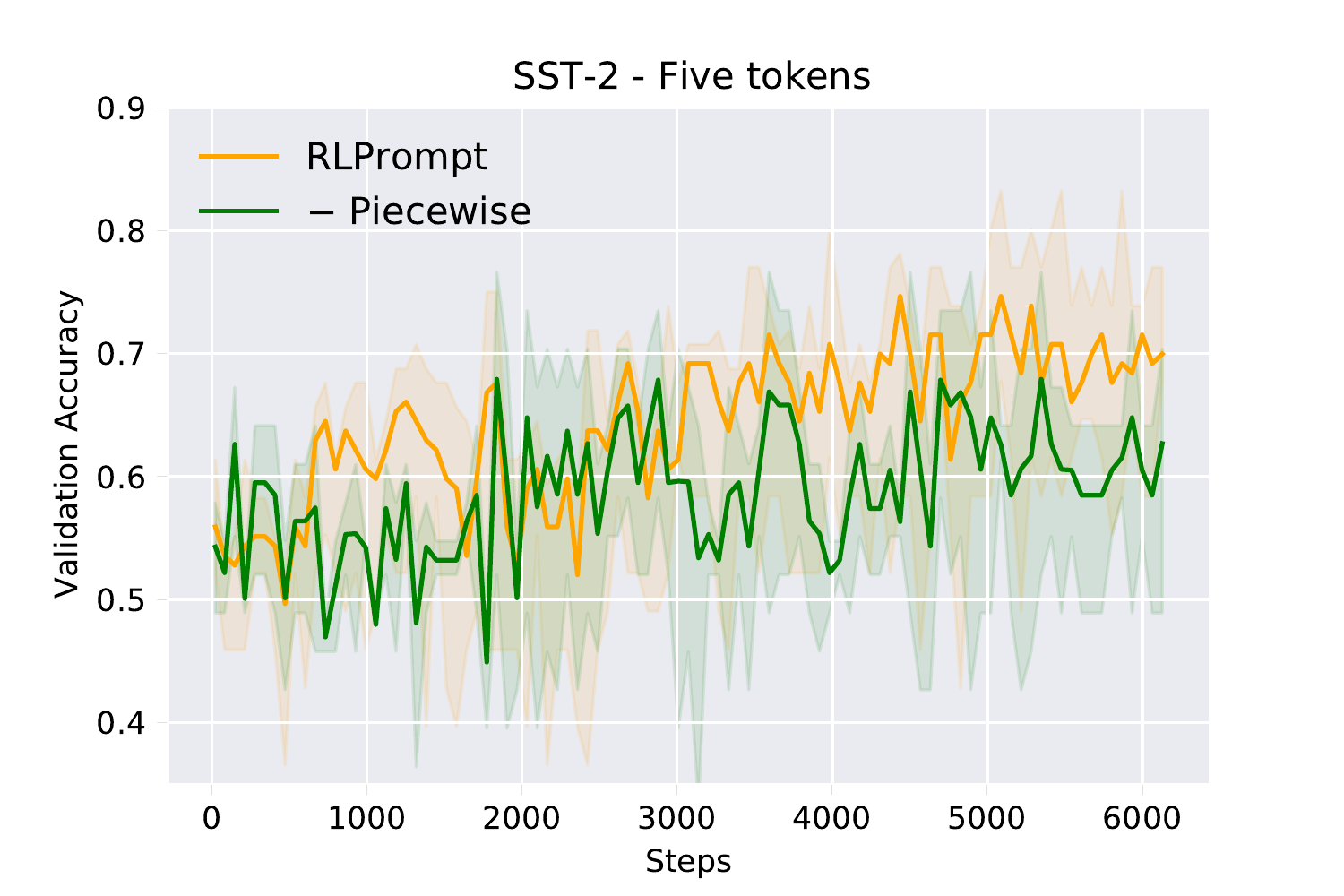}
    \includegraphics[width=0.48\textwidth]{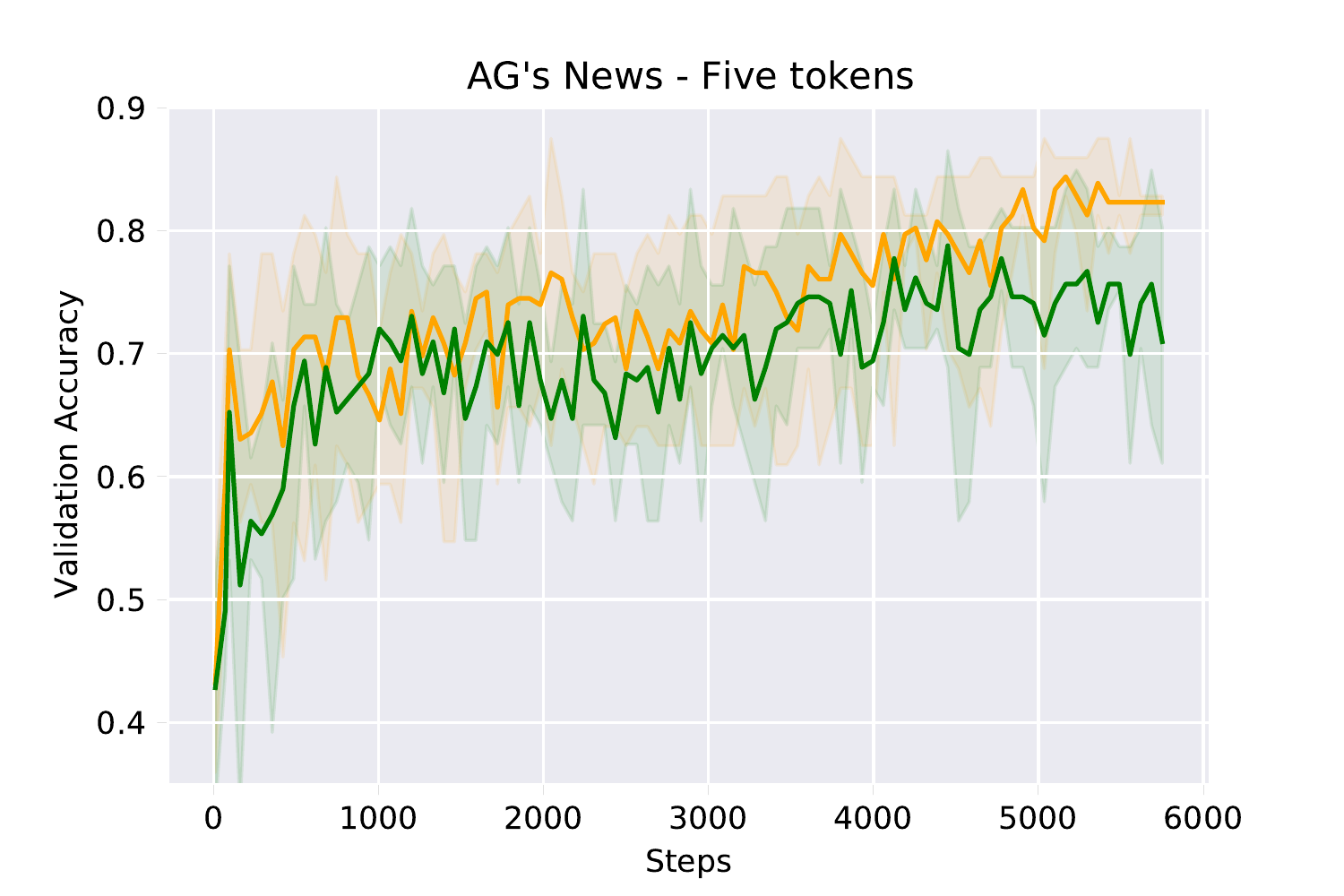}
    \vspace{-10pt}
    \caption{Comparison of our method with ({\color{orange} orange}) and without ({\color{teal} green}) piecewise reward function for few-shot classification. The format is the same as Figure~\ref{fig:train-efficiency}.
    }
    \label{fig:sst2-ablations}
\end{figure}

\input{tab-NLU-dataset}

\input{tab-NLU-additional-result}

\label{appendix:implementation:classify}
\paragraph{Additional Training Details}
During training, we explore the prompt space using top-256 sampling from the policy network, whose input is just one placeholder word ``classification''. 
To update the parameters, we use an Adam~\cite{kingma2014adam} optimizer with learning rate $5 \times 10^{-5}$. 
Furthermore, we multiply all rewards by 5 to increase the reward scale of well-performing prompts, and apply $z$-score normalization (\S\ref{reward: z-score}) across prompts for more efficient learning.
We train the policy with 16 prompts per batch for 6K steps for 2 tokens, 12k steps for 5 tokens, and compute validation performance every 10 steps.
Using an NVIDIA GeForce RTX 3090 GPU, each experiment typically takes from 1.5 hours using distilRoBERTa-base to 4 hours using RoBERTa-large.
During evaluation, we average the performance of 3 prompts with the highest validation accuracy for each experiment. 
Due to the instability and inherent randomness of the few-shot setup~\cite{henderson2018RLrandom, gao2021LMBFF},
we sample $5$ different training and validation sets, run $3$ experiments per set with different random seeds, and report the average accuracy and standard deviation.

\paragraph{Additional Results} 
We present our results on the additional datasets described in Section~\S\ref{subsec:classifysetting} in Table~\ref{tab:cls-addition}. Again, our method outperforms prompting baselines on average. Methods tuning continuous parameters such as Fine-Tuning, Prompt Tuning, and BB Tuning show better performance on Yahoo and DBPedia, both multi-way datasets which have much more training data under our setting (e.g., Yahoo with 16 classes has 256 training examples, whereas SST-2 with 2 classes has only 32 examples). Expensively updating all parameters, Fine-Tuning achieves the highest average accuracy on these larger datasets.

\paragraph{Ablation Study} 
\label{appendix:ablation}
As mentioned before (\S\ref{subsec:method:reward-engineering}), misspecified or vulnerable reward functions can prevent the policy from discovering truly strong-performing prompts.
To address this challenge, we propose to design piecewise reward functions 
that provide bonus to qualitative behaviors such as achieving certain accuracies on each class.
As our reward function for few-shot classification adopts this design, we assess its effectiveness by ablating the piecewise component.
Specifically, we test on
SST-2 \cite{socher2013recursive} and AG's News \cite{zhang2015character} using 5 prompt tokens with the distilRoBERTa-base model as an example. 
We run 5 RL experiments on the same few-shot dataset using different random seeds, and compute the validation accuracy every 50 steps. 
As the results in Figure~\ref{fig:sst2-ablations} show, our piecewise reward function 
improves training stability by leading to strong-performing prompts more consistently, resulting in better average performance across random seeds and datasets.

\begin{table*}[t]
\centering
\small

{\renewcommand{\arraystretch}{1}
\setlength{\tabcolsep}{5pt}
\begin{tabular}{llllll | lll}
\toprule
 & & & & & & \multicolumn{2}{c}{Content Preservation} & Fluency \\
{Model}    & {Content}    & {Style}      & {Fluency}    & {\bf $\bm{J}$({\scriptsize C, S, F})} & {\bf GM({\scriptsize C, S, F})} & {BLEU}       & {BERTScore}  & {PPL}$\downarrow$        \\ 
\midrule
\rowcolor{Gray}
\multicolumn{9}{l}{\textit{Training Baselines (Full Data)}}                                                                             \\
Deep Latent  & 47.1                & \textbf{70.8} & 49.8                & 17.8                & 55.0                 & \textbf{19.2} & 38.3          & 78.2          \\
STRAP        & 54.6                & 69.3          & 85.0                  & \textbf{30.3}       & \textbf{68.5}        & 16.3          & \textbf{46.3} & \textbf{33.3} \\
\rowcolor{Gray}
\multicolumn{9}{l}{\textit{Prompting Baselines (GPT-2-xl)}}                                           \\
Null Prompt        & 41.9 \scriptnumber{2.4}          & 56.1 \scriptnumber{5.0}    & 87.6 \scriptnumber{1.1}          & 17.3 \scriptnumber{1.2}          & 59.0 \scriptnumber{0.8}           & 9.3 \scriptnumber{0.8}     & 32.7 \scriptnumber{1.0}    & 48.1 \scriptnumber{1.4}    \\
Random Prompt      & 46.8 \scriptnumber{2.6}          & 55.0 \scriptnumber{4.7}    & \textbf{89.4 \scriptnumber{0.8}} & 17.7 \scriptnumber{1.3}          & 61.2 \scriptnumber{0.8}           & 10.9 \scriptnumber{0.7}    & 34.8 \scriptnumber{1.0}    & 50.5 \scriptnumber{1.6}    \\
Manual Prompt      & \textbf{58.8 \scriptnumber{2.7}} & 52.9 \scriptnumber{4.5}    & 82.2 \scriptnumber{1.7}          & 22.2 \scriptnumber{1.9}          & 63.4 \scriptnumber{1.5}           & 14.0 \scriptnumber{0.7}    & 40.4 \scriptnumber{0.7}    & 62.4 \scriptnumber{1.5}    \\
\rowcolor{Gray}
\multicolumn{9}{l}{\textbf{\textit{\modelname (Ours -- 100-Shot)}}}                                                                                       \\
GPT-2-xl & 51.8 \scriptnumber{1.5}          & 65.1 \scriptnumber{2.7}    & 85.2 \scriptnumber{0.3}          & \underline{26.7} \scriptnumber{1.3}          & \underline{66.0} \scriptnumber{0.9}           & 13.1 \scriptnumber{0.4}    & 39.0 \scriptnumber{0.8}    & 63.2 \scriptnumber{1.3}   \\ \bottomrule
\end{tabular}
}
\caption{ Automatic evaluation of our method vs. baselines on the Shakespeare \cite{xu-etal-2012-shakespeare} authorship transfer dataset. For this dataset, our method only uses 100 examples per style, and numbers in (parentheses) are standard deviations across 3 randomly-drawn training sets. The metrics are the same as Tables \ref{tab:tst-yelp-auto} and \ref{tab:tst-yelp-addl}.}
\label{tab:tst-shakespeare-auto}
\end{table*}

\begin{table}[h]
\centering
\vspace{6pt}
{\renewcommand{\arraystretch}{1.0}
\setlength{\tabcolsep}{5pt}
\small
\begin{tabular}{llll}
\toprule
& \multicolumn{2}{c}{Content Preservation} & Fluency \\
{Model}    &  {BLEU}       & {BERTScore}  & {PPL}$\downarrow$        \\ 
\midrule
\rowcolor{Gray}
\multicolumn{4}{l}{\textit{Training Baselines}}                                                                                                                                                             \\ 
Style Transformer &  27.6           & 56.1           & 78.2           \\
DiRR              & \textbf{30.0} & \textbf{61.7} & 40.6           \\ 
\rowcolor{Gray}
\multicolumn{4}{l}{\textit{Prompting Baselines (GPT-2-xl)}}                                                                                                                                             \\ 
Null Prompt       &  6.6            & 35.8           & 59.5           \\
Random Prompt     &  7.3            & 37.4           & 60.5           \\
Manual Prompt     &  19.2 \scriptnumber{4.1}          & 53.1 \scriptnumber{5.0}          & 35.5 \scriptnumber{9.0}          \\ 
\rowcolor{Gray}
\multicolumn{4}{l}{\textbf{\textit{\modelname (Ours)}}}                                                                                                                                                                    \\ 
distilGPT-2       & 15.7 \scriptnumber{0.7}          & 49.1 \scriptnumber{0.6}          & 43.6 \scriptnumber{0.6}          \\
GPT-2-small       & 16.5 \scriptnumber{0.4}          & 51.3 \scriptnumber{0.6}          & 37.8 \scriptnumber{4.8}          \\
GPT-2-medium      & 20.0 \scriptnumber{1.2}          & 55.1 \scriptnumber{1.1}          & 34.4 \scriptnumber{0.8}          \\
GPT-2-large       & 19.8 \scriptnumber{0.5}          & 54.7 \scriptnumber{0.7}          & 34.9 \scriptnumber{1.4}          \\
GPT-2-xl      & 24.2 \scriptnumber{1.2}          & 59.0 \scriptnumber{0.8}          & \textbf{34.3 \scriptnumber{0.9}} \\ 
 \bottomrule
\end{tabular}
}
\caption{ 
Additional automatic evaluation results on Yelp \cite{shen2017style} sentiment transfer.
BLEU and BERTScore are computed between outputs and references. PPL is the perplexity under a GPT-2 language model. 
Numbers in (parentheses) are standard deviations across 3 sets of prompts. 
}
\label{tab:tst-yelp-addl}
\end{table}

\subsection{Text Style Transfer}
\label{appendix:implementation:tst}

\paragraph{Reward Function Details}
We implement our content preservation reward using its CTC metric \cite{deng-etal-2021-compression}, which measures the bi-directional information alignment between input $\xv$ and output $\yv$. We compute the alignment by matching token embeddings from RoBERTa-large similarly to BERTScore \cite{zhang2019bertscore}, a technique that shows the highest correlation with human judgments. For the style reward, we compute the target style probability under a BERT-base-uncased classifier learned from the training data. 

\paragraph{Dataset Statistics} \textbf{(1)} Yelp~\cite{shen2017style} contains 266K positive and 177K negative reviews for training, 38K and 25K for validation, and 76K and 50K for testing, respectively. We perform evaluation on a separate dataset consisting of 500 reviews for each sentiment, with reference outputs collected by \citet{li-etal-2018-delete}. 
\textbf{(2)} We use the Shakespeare \cite{xu-etal-2012-shakespeare} dataset compiled by \citet{jhamtani-etal-2017-shakespearizing}, which contains 18K parallel sentence pairs from Shakespeare's plays and their modern translations for training, 1.2K for validation, and 1.4K for testing. We treat the dataset as a non-parallel corpus for training, but use the paired sentences as reference during evaluation. We preprocess both datasets with a simple text cleaning function to remove tokenization artifacts (e.g., ``\texttt{it 's great .}'' becomes ``\texttt{it's great.}''). We include the function in our public codebase for reproducibility.

\begin{figure}[t]
    \centering
    \includegraphics[width=0.48\textwidth]{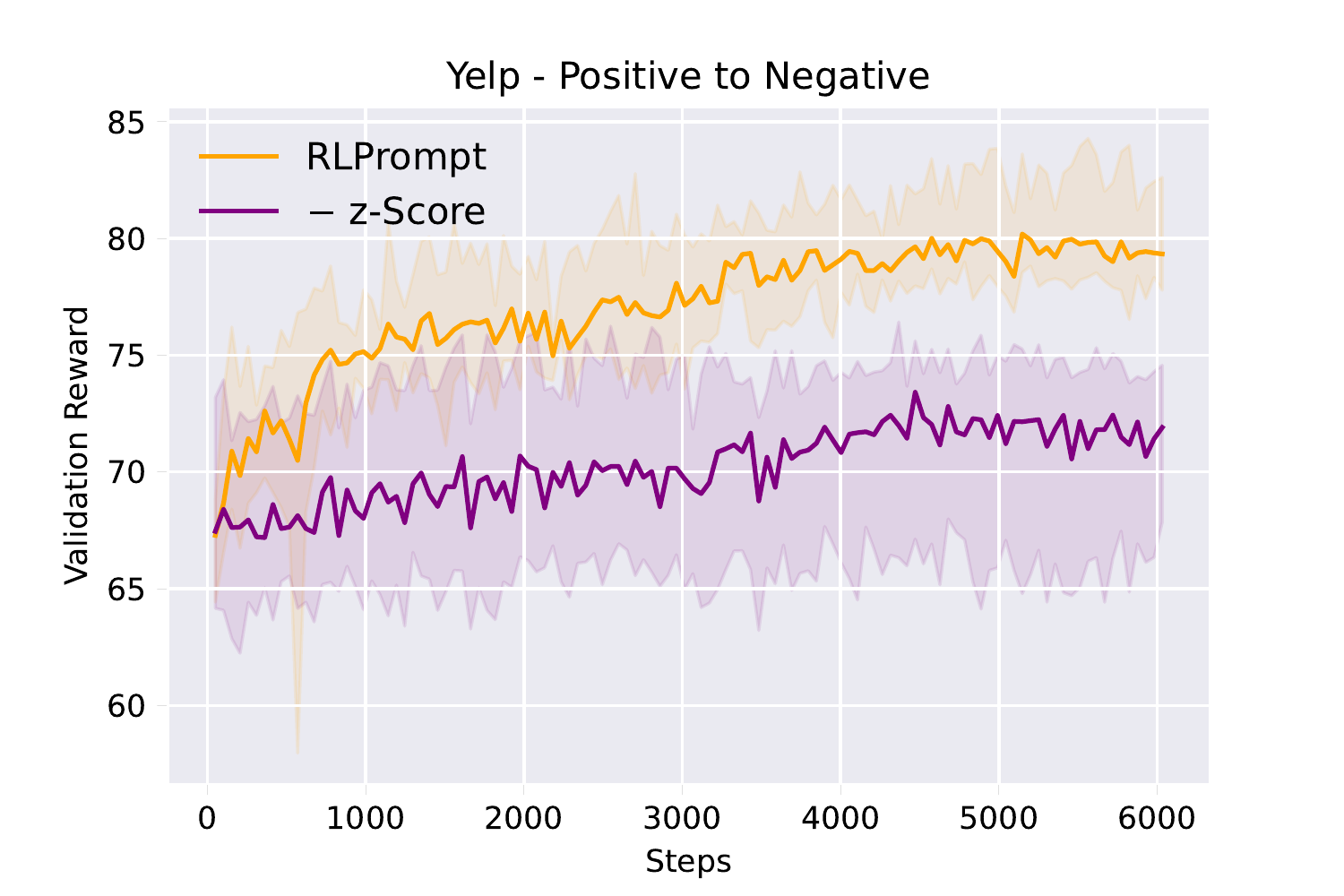}\\
    \vspace{-10pt}
    \caption{Additional comparison of our method with ({\color{orange} orange}) and without ({\color{purple} purple}) $z$-score reward normalization. The format is the same as Figure~\ref{fig:train-efficiency}.}
    \label{fig:yelp-ablations-2}
\end{figure}

\paragraph{Additional Training Details}
In training, we sample 4 prompts for each input using top-50 sampling from our policy network. During sampling, we bias all logits by -10 to encourage exploration. 
For each prompt, we generate outputs using top-10 sampling,
and bootstrap the reward 4 times to reduce variance. For SQL training, we set the target learning rate to be $10^{-3}$, and shape the reward from a scale of [0,1] to [-20,80]. We optimize the prompt generator using an Adam optimizer with learning rate $10^{-4}$, except for Yelp negative-to-positive and Shakespeare using GPT-2-large and GPT-2-xl models, which we train with learning rate $5 \times 10^{-5}$.
We train 2 inputs per batch for 6K steps if learning rate is $10^{-4}$, and 12K steps if the learning rate is $5 \times 10^{-5}$.
Also using the RTX 3090 GPU, each experiment typically takes from 10 hours using distilGPT-2 to 1 day using GPT-2-xl.
To reduce the performance variance caused by sample selection and RL initialization, we average the performance from 5 evaluation runs for each of 3 RL experiments using our own method. 
Additionally, we perform the same sample selection for all our baselines for comparable performance. 
For Shakespeare training baselines, we do not perform sample selection in order to avoid biasing the full-dataset models with our few-shot style classifiers. 

\paragraph{Evaluation Details}
For automatic evaluation, We measure Content using
the CTC metric \cite{deng-etal-2021-compression} discussed earlier.
To compute Style, we train BERT-base-uncased classifiers on both training and testing data, with validation accuracies of 98.4\% and 93.7\% on Yelp and Shakespeare, respectively.
To evaluate Fluency, we rate output grammaticality using the classifier from \citet{krishna-etal-2020-reformulating}.\footnote{\url{https://huggingface.co/cointegrated/roberta-large-cola-krishna2020}}
We also report popular metrics such as 
BLEU~\citep[using sacreBLEU,][]{post-2018-call} and BERTScore~\cite{zhang2019bertscore} for content preservation, 
and perplexity (PPL) for fluency.
To compute PPL, we fine-tune GPT-2 LMs on each TST dataset.
For human evaluation, we enlist 5 graduate students who are fluent in English to rate Content, Style, and Fluency on a Likert scale of 1-5, and collect 3 ratings for each output. 
The average inter-rater agreement is 0.35 in terms of Fleiss' kappa \cite{fleiss1973equivalence}, which is fair and similar to previous work \cite{mir-etal-2019-evaluating}.


\paragraph{Few-Shot Experiment Details}
As discussed before, we experiment with few-shot text style transfer on the Shakespeare dataset. For the training baselines, we compare with Deep Latent \cite{he2020probabilistic} and STRAP \cite{krishna-etal-2020-reformulating}, both trained on the full data.  
STRAP fine-tunes a GPT-2 \cite{radford2019GPT2} with self-supervised paraphrasing signals, which can be seen as a full-model tuning analogue to our method. We also compare with the same prompting baselines tested for Yelp. 
Both prompting baselines and our method use GPT-2-xl as the task LM.

\paragraph{Few-Shot Experiment Results}
We present the automatic evaluation results for Shakespeare in Table~\ref{tab:tst-shakespeare-auto} to illustrate our few-shot performance. 
Even with only 100 training examples and no update to the model, our method outperforms or gets close to training baselines using the full dataset such as Deep Latent and STRAP. STRAP is also limited to a subset of styles (e.g., authorship and formality), whereas our method accommodates a wider range of styles. Compared to prompting baselines, our method not only improves the performance, but also shows higher robustness to randomly-drawn training sets, as evidenced by the lower standard deviations for Content and Style.

\section{Additional Analysis}
\label{appendix:additional-analysis}

\paragraph{Fluent vs. Gibberish Prompts}
We propose to optimize fluent prompts with top-k filtering \citep{qin2022cold}. That is, we limit our policy's action space at each step $t$ to the tokens with top-20 probabilities under a GPT-2 LM, conditioning on the previous prompt tokens $\zv_{<t}$.
Other than that, we train the policy using the same routine.
To evaluate prompt perplexity, we use an out-of-the-box GPT-2 model.

\begin{figure}[t]
    \centering
    \includegraphics[width=0.48\textwidth]{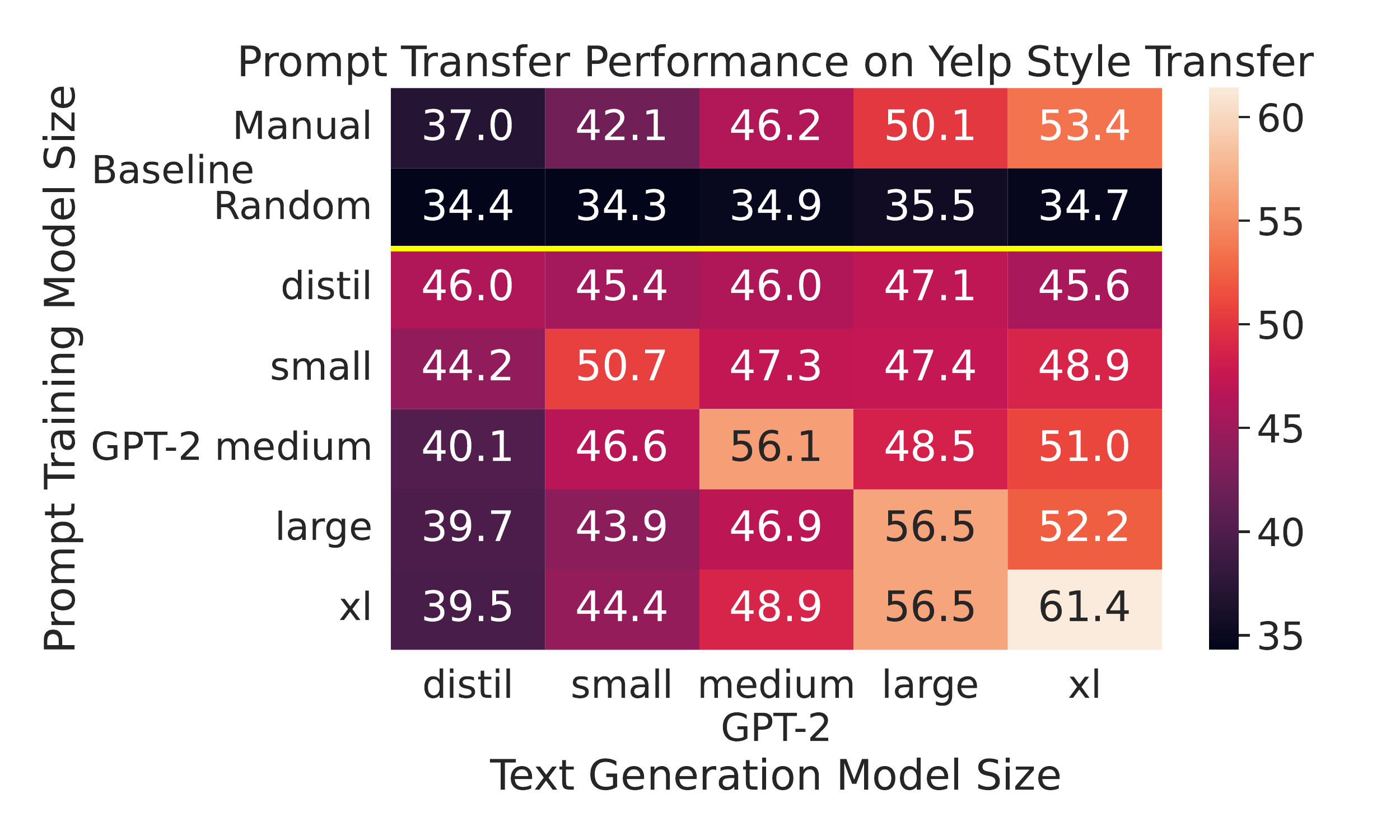}
    \\
    \vspace{-10pt}
    \caption{ Heatmap of Yelp style transfer performance with transferred discrete prompts. The columns represent the models used to learn the prompts, and the rows represent the models we perform text generation with. \texttt{Manual} and \texttt{Random} refer to the baselines presented in Table \ref{tab:tst-yelp-auto}. Brighter color represents better joint score $J(\cdot)$. }
    \label{fig:prompt-transfer}
    \vspace{10pt}
\end{figure}

\paragraph{Transferring Prompts across LMs}
Previously, we presented our prompt transfer results for few-shot classification in Section~\S\ref{subsec:analysis}. For text style transfer, We use the prompts trained for each size of GPT-2 (from the smallest distil to the largest xl) to perform generation using every other model, and present the average performance over 5 evaluations in the heatmap of Figure~\ref{fig:prompt-transfer}.
We also include Manual Prompt for comparison and Random Prompt for the baseline performance without transfer.
Manual Prompt shows uniformly worse performance than learned prompts with smaller models like distilGPT-2 and GPT-2-small, but generally better results with larger models like GPT-2-large and -xl, suggesting that human-written prompts may better activate larger models. 
Overall, all optimized prompts see some transfer, as evidenced by the uniformly better performance than Random Prompt, and the level of success depends on both the prompt training and generation models, similarly to classification.

\paragraph{Qualitative Analysis of Prompt Tokens}
Empowered by the transparency of discrete tokens, we investigate the prompts we learned for classification to characterize the similar patterns learned by different LMs discovered by the prompt trasfer analysis (\S\ref{sec: Transfer across Models}).
In particular, we frequently find semantically similar tokens among our learned prompts, which we name ``strong words'' and list in Table \ref{tab:strong words}. 
These strong words make sense in the context of their specific tasks, indicating the LMs may indeed capture certain human-understandable patterns during pre-training. 
For instance, ``absolutely'' may signal strong opinion before judging a sentence as positive or negative, whereas ``News'' appears to be a hint for classifying the topic of a news piece. 
Besides these semantically meaningful prompt tokens, we also find some unintelligible prompts that nevertheless achieve good performance on downstream tasks, or so-called ``secret language~\cite{Daras2022DiscoveringTH} of the LM'' (e.g., ``\texttt{imentariesariesaryary}'' can reach 80\% accuracy with RoBERTa-large on AG's News).


\begin{table}[t]
    \small
    \resizebox{\columnwidth}{!}{
    \begin{tabular}{@{}lcc@{}}
    \toprule
    Verbalizers & \modelname & Manual \\ 
    \midrule
    World, Sports, Business, Tech & \textbf{77.6} \scriptnumber{1.5} & 76.9 \\
    Global, Athletics, Finance, Technology & \textbf{65.3} \scriptnumber{0.5} &  63.5 \\
    \bottomrule
    \end{tabular}}
    \caption{ Comparison of our method vs. Manual Prompt on AG's News using different verbalizers. The manual prompt is ``News:" and our prompts consist of 2 tokens.}
    \label{tab:verbalizers2}
\end{table}

Beyond finding strong words, we also study whether we can construct strong-performing prompts by arbitrarily composing these strong words, which can provide insight into whether LMs use these strong words compositionally.
To this end, we construct several prompts, evaluate their downstream performance, and tabulate the results in Table~\ref{tab:strongprompt}. 
Interestingly, composing more strong words indeed can lead to improved performance, but the level of success is sensitive to various factors, such as word order and the specific tokens we choose, indicating that existing LMs are still brittle even when responding to discrete tokens learned from optimization.


\begin{table}[t]
    \centering
{\renewcommand{\arraystretch}{1.2}
\small
    \begin{tabular}{@{}rp{4cm}@{}}
    \toprule
    Task Category & Strong Words \\ 
    \midrule
    Sentiment Analysis & \textbf{Absolutely}, \textbf{absolutely}, \textbf{Totally}, \textbf{downright}, profoundly, VERY, Very, Really, highly \\ 
    \begin{tabular}[t]{@{}r@{}} News Classification \end{tabular}
    &
    \textbf{News},
    \textbf{Reviewer}, 
    \textbf{Reports},
    \textbf{reported}, \textbf{Staff}, \textbf{Information}, \textbf{Statement}, Stories, Guide, say, \\
    \bottomrule
    \end{tabular}}
    \caption{Strong words from \modelname for different task categories. The words are all sensitive to cases and to whether we prepend the special character Ġ.}
    \label{tab:strong words}
    \vspace{8pt}
\end{table}

\begin{table}[t]
    \small
{\renewcommand{\arraystretch}{1}
    \resizebox{\columnwidth}{!}{
    \begin{tabular}{lcc}
    \toprule
    Template & RoBERTa & GPT-2 \\ 
    \midrule
    \rowcolor{Gray} \multicolumn{3}{l}{SST-2} \\
    \texttt{\textless{}S\textgreater{}} downright \texttt{[MASK]} . &  80.6 &  86.7 \\
    \texttt{\textless{}S\textgreater{}} Really downright \texttt{[MASK]} . &  90.4 &   89.1\\
    \texttt{\textless{}S\textgreater{}} Absolutely \texttt{[MASK]} . & 91.7 & 87.8\\
    \texttt{\textless{}S\textgreater{}} AbsolutelyAbsolutely \texttt{[MASK]} . & 89.2 & 72.3\\
    \begin{tabular}[c]{@{}l@{}} \texttt{\textless{}S\textgreater{}} Absolutely VERY absolute \\   VERY absolute \texttt{[MASK]} . \end{tabular}
    &  92.7 &  73.8 \\
    \midrule
    \rowcolor{Gray} \multicolumn{3}{l}{AG's News} \\
    \texttt{[MASK]} Reviewer \texttt{\textless{}S\textgreater{}} & 74.5 & --- \\
    \texttt{[MASK]} Reviewer Stories \texttt{\textless{}S\textgreater{}} & 81.0 & --- \\
    \texttt{[MASK]} StaffInformationStatement \texttt{\textless{}S\textgreater{}} &
    76.8 & --- \\
    \begin{tabular}[c]{@{}l@{}} \texttt{[MASK]} StaffInformationStatement \\  Reviewer Stories \texttt{\textless{}S\textgreater{}} \end{tabular}
    &
    79.8 & --- \\
    \bottomrule
    \end{tabular}}
    }
    \caption{The performance of manual prompt examples by composing strong words from Table~\ref{tab:strong words} for both sentiment analysis and news topic classification across RoBERTa-large and GPT-2-large.}
    \label{tab:strongprompt}
\end{table}

\section{Additional Related Work}
\label{appendix:related-work}

\subsection{Prompting Paradigms}
\paragraph{Fine-Tuning} 

The conventional approach to using pre-trained LMs is fine-tuning model parameters on downstream datasets \cite{devlin-etal-2019-bert, liu2019roberta, lewis2020bart, raffel2020T5, radford2019GPT2}. 
While driving progress in a wide range of NLP tasks, fine-tuning expensively updates all model parameters and shows limited success with small datasets. 
Prompt-based fine-tuning \cite{gao2021LMBFF,schick-schutze-2021-just} uses prompting to improve few-shot performance, but the problem of costly training remains unsolved.  

\paragraph{Manual Prompt} 

As LMs show remarkable progress in understanding natural language \cite{peters-etal-2018-deep,devlin-etal-2019-bert}, 
researchers first use hand-crafted fill-in-the-blank prompts to extract knowledge from pre-trained LMs for probing analyses \cite{petroni2019KB,jiang2020can}. Later on, \citet{brown2020language} show that using manually-written prompts, large LMs can perform a number of NLU and NLG tasks without any training examples.
Meanwhile, other studies \cite{raffel2020T5,schick2021exploiting,sanh2021T0} formulate various NLP tasks as manual prompts. 

\paragraph{Instructions} 

Separate from but related to manual prompts, another line of work \cite{weller2020learning,efrat2020turking,mishra2021NI,wang2022NI2} makes use of instructional prompts which provide task descriptions instead of fill-in-the-blank questions. 
In particular, instruction meta-tuning \cite{mishra2021NI,zhong-etal-2021-adapting-language,wei2022finetuned} trains models on some tasks with instructions and supervised data in order to generalize to unseen tasks formulated as instructions without training examples. 

\paragraph{In-Context Demonstration} 
Besides zero-shot learning, \citet{brown2020language} achieve more remarkable performance on few-shot learning by inserting training examples into the input context. More recent works \cite{gao2021LMBFF,liu2021KATE,lu2021fantastically,min2022rethinking} further explore the selection and analysis of in-context demonstrations.
\citet{reif2021recipe} propose augmented zero-shot learning, which inserts training examples from related tasks as demonstrations for tasks without supervised training data, such as text style transfer.

\paragraph{Discrete Prompt Enumeration} 

Because discrete prompts are difficult to optimize and susceptible to small design variations \cite{zhao2021calibrate, webson2021prompt, lu2021fantastically}, a number of existing works seek to locate better prompts by augmenting human-written prompts with heuristics such as paraphrasing \cite{jiang2020can, gao2021LMBFF}, editing \cite{prasad2022grips}, and reframing \cite{mishra2021reframing}. The final prompt is typically selected to maximize some downstream performance metric. 

\paragraph{AutoPrompt} 

\citet{shin2020autoprompt} optimize discrete prompts by editing prompt tokens with guidance from model gradients. 
While seeing some success with large training data, the method relies heavily on approximation, which leads to less stable training and limited applicability to few-shot settings. 

\paragraph{Soft Prompt Tuning}

Replacing discrete prompts with continuous embeddings, several parallel works \cite{qin-eisner-2021-learning,li2021prefix,liu2021ptuningv1} propose to optimize soft prompts with gradient-based tuning. Soft prompt tuning can be seen as a variant of parameter-efficient transfer learning \cite{houlsby2019parameter, he2021towardsunified, ding2022delta}, and inspires a number of follow-up works that boost its performance \citep[e.g.,][]{liu2021ptuningv2, gu2021ppt, vu2021spot, clive2021control} or explore novel applications \citep[e.g.,][]{tan-etal-2022-msp,zhou2022conditional,levine2022standing}. 
By its nature, however, soft prompts are difficult for humans to understand because of its continuous form \cite{khashabi2021prompt,lester2021promptuning,hambardzumyan2021warp,mokady2021clipcap}. Defined in the latent space of specific models, learned prompts are also virtually impossible to use with a different model. Furthermore, their training typically requires gradient information from the models they prompt, which can be expensive to compute or simply inaccessible for models deployed as inference API, such as GPT-3 \cite{brown2020language}.
\citet{sun2022black} and \citet{diao2022black} propose black-box tuning, which updates continuous prompts using gradient-free techniques to some success. 

\subsection{Controllable Text Generation}
Current state-of-the-art models typically fine-tune entire pre-trained LMs 
\citep[e.g.,][]{ziegler2019fine, keskar2019ctrl, ziegler2019encoder,liu2021DIRR}. 
Recent work instead employs various prompts to steer the LM to generate text with properties such as topic \cite{guo2021text, qian2022contrastiveprefix} and (lack of) toxicity \cite{liu2021dexperts,perez2022red}, or from modalities such as image \cite{mokady2021clipcap,zhou2022conditional}, structured data \cite{li2021prefix, an2022input}, and numbers \cite{wei2022chain}. 
However, these works either control simple attributes, perform no explicit prompt optimization, or have access to supervised data. 
For unsupervised tasks with more complex requirements such as text style transfer \cite{hu2017toward, jin2022deep}, \citet{reif2021recipe} proposed augmented zero-shot prompting, 
which achieves some success using huge LMs (e.g., GPT-3). 
Complementary to the works above which focus on finding prompts, \citet{zou2021controllable} augment the generation decoding objective using the prompt, leading to improved performance in poetry generation and long-form QA.

\input{tab-NLU-instruction}

\begin{table*}[h]
\setlength{\tabcolsep}{2.5pt}
\centering
{\renewcommand{\arraystretch}{1.2}
\small
\begin{tabular}{llrrrrrrrr} \toprule
ID                                                  
& {\begin{tabular}[c]{@{}l@{}}Template \\ {[}\textcolor{red}{to negative} | {\textcolor{teal}{to positive}}{]}\end{tabular}}                                                                                                            & \multicolumn{1}{l}{{Content}} & \multicolumn{1}{l}{{Style}} & \multicolumn{1}{l}{{Fluency}} & \multicolumn{1}{l}{{\bf $\bm{J}$({\scriptsize C, S, F})}} & \multicolumn{1}{l}{\textbf{GM({\scriptsize C, S, F})}} & \multicolumn{1}{l}{{BLEU}} & \multicolumn{1}{l}{{BERTScore}} & \multicolumn{1}{l}{{PPL$\downarrow$}} \\ \midrule
\rowcolor{Gray}
\multicolumn{10}{l}{\textit{Null Prompt}}                   \\
1                                                 
& "\{input\}" "                                                                                                                                                                                           & 37.4 \scriptnumber{0.1}                                & 94.8 \scriptnumber{0.1}                             & \textbf{97.6 \scriptnumber{0.1}}                        & 33.6 \scriptnumber{0.1}                                  & 70.2 \scriptnumber{0.1}                                   & 6.6 \scriptnumber{0.1}                              & 35.8 \scriptnumber{0.1}                                 & 59.5 \scriptnumber{2.0}                            \\ 
\vspace{0cm} \\
\rowcolor{Gray}
\multicolumn{10}{l}{\textit{Manual Prompt}}                   \\
 1                                            
& \begin{tabular}[c]{@{}l@{}}Here is some text: "\{input\}". \\ Here is a rewrite of the text, \\ which is more \\ {[}\textcolor{red}{negative} | \textcolor{teal}{positive}{]}: "\end{tabular}                                              & 72.1 \scriptnumber{0.1}                               & 94.8 \scriptnumber{0.3}                             & 91.6 \scriptnumber{0.1}                               & 62.3 \scriptnumber{0.2}                                  & \textbf{85.6 \scriptnumber{0.1}}                            & 23.9 \scriptnumber{0.1}                             & 58.8 \scriptnumber{0.1}                                 & \textbf{29.6 \scriptnumber{0.3}}                 \\ 
\vspace{0cm} \\
 2                                              & \begin{tabular}[c]{@{}l@{}}Change the following sentence \\ from {[}\textcolor{red}{positive} | \textcolor{teal}{negative}{]} \\ sentiment to {[}\textcolor{red}{negative} | \textcolor{teal}{positive}{]} \\ sentiment but keep its \\ semantics. "\{input\}" "\end{tabular} & 60.4 \scriptnumber{0.1}                              & 91.9 \scriptnumber{0.2}                             & 94.0 \scriptnumber{0.1}                                & 50.5 \scriptnumber{0.1}                                  & 80.5 \scriptnumber{0.1}                                   & 17.4 \scriptnumber{0.1}                            & 51.3 \scriptnumber{0.1}                                & 31.0 \scriptnumber{0.4}                           \\ 
 \vspace{0cm} \\
 3                                              & \begin{tabular}[c]{@{}l@{}}"\{input\}". Rewrite the sentence \\ to be {[}\textcolor{red}{sadder} | \textcolor{teal}{happier}{]} but \\ have the same meaning. "\end{tabular}                                                               & 60.2 \scriptnumber{0.2}                                & 87.7 \scriptnumber{0.4}                              & 94.0 \scriptnumber{0.2}                                & 47.4 \scriptnumber{0.3}                                   & 79.2 \scriptnumber{0.1}                                    & 16.2 \scriptnumber{0.1}                             & 49.3 \scriptnumber{0.1}                                  & 45.8 \scriptnumber{0.7}                            \\ 
 \vspace{0cm} \\
 \rowcolor{Gray}
\multicolumn{10}{l}{\textit{Fluent Prompt}}                   \\
 1                                              & \begin{tabular}[c]{@{}l@{}}{[}\textcolor{red}{I don't like having} | \\ \textcolor{teal}{I love my life (}] "\{input\}" "\end{tabular}                             & 54.1 \scriptnumber{0.5}                                & 95.2 \scriptnumber{0.4}                               & 93.9 \scriptnumber{0.7}                                & 47.4 \scriptnumber{0.4}                                   & 78.5 \scriptnumber{0.3}                                    & 13.4 \scriptnumber{0.4}                             & 45.7 \scriptnumber{0.2}                                  & 52.3 \scriptnumber{1.9}                            \\ 
 \vspace{0cm} \\
 2                                              & \begin{tabular}[c]{@{}l@{}}{[}\textcolor{red}{\ This is not an example} |\\ \textcolor{teal}{The best is good\textbackslash{}n}{]} "\{input\}" "\end{tabular}                & 51.5 \scriptnumber{0.1}                                & \textbf{96.8 \scriptnumber{0.4}}                              & 94.2 \scriptnumber{0.6}                                & 46.0 \scriptnumber{0.4}                                   & 77.7 \scriptnumber{0.1}                                    & 11.9 \scriptnumber{0.3}                             & 46.2 \scriptnumber{0.2}                                  & 35.4 \scriptnumber{2.3}                            \\ 
 \vspace{0cm} \\
 3                                              & \begin{tabular}[c]{@{}l@{}}{[}\textcolor{red}{I don't like} |\\ \textcolor{teal}{I love my work (}{]} "\{input\}" "\end{tabular}                & 51.5 \scriptnumber{0.4}                                & 96.6 \scriptnumber{0.7}                              & 95.7 \scriptnumber{0.5}                                & 46.7 \scriptnumber{0.5}                                   & 78.1 \scriptnumber{0.2}                                    & 12.3 \scriptnumber{0.3}                             & 46.2 \scriptnumber{0.3}                                  & 43.5 \scriptnumber{1.3}                            \\ 
 \vspace{0cm} \\
\rowcolor{Gray}
\multicolumn{10}{l}{\textbf{\textit{\modelname (Ours)}}}                   \\
1
& \begin{tabular}[c]{@{}l@{}}{[}\textcolor{red}{Fixed ($-$ contrasts ($-$ contrasts} | \\ \textcolor{teal}{Dutch English excellent Correct} \\ \textcolor{teal}{(\textgreater{}}{]} "\{input\}" "\end{tabular}                                                     & 71.5 \scriptnumber{0.1}                                & {96.6 \scriptnumber{0.2}}                      & 90.1 \scriptnumber{0.2}                                & \textbf{62.8 \scriptnumber{0.9}}                           & 85.4 \scriptnumber{0.1}                                    & 23.5 \scriptnumber{0.1}                             & 58.7 \scriptnumber{0.1}                                  & 34.1 \scriptnumber{0.2}                            \\ 
\vspace{0cm} \\
2
& \begin{tabular}[c]{@{}l@{}}{[}\textcolor{red}{Fixed RemovedChanged} \\ \textcolor{red}{Prevent outcomes} | \\ \textcolor{teal}{Parameters Comparison}\\ \textcolor{teal}{)=( Compare either}{]} \\ "\{input\}" "\end{tabular}                                                 & 71.0 \scriptnumber{0.1}                                & 91.9 \scriptnumber{0.3}                              & 89.3 \scriptnumber{0.2}                                & 58.9 \scriptnumber{1.1}                                   & 83.5 \scriptnumber{0.1}                                    & 23.7 \scriptnumber{0.1}                             & 58.3 \scriptnumber{0.1}                                  & 35.3 \scriptnumber{0.5}                            \\
\vspace{0cm} \\ 
3
& \begin{tabular}[c]{@{}l@{}}{[}\textcolor{red}{Affect differed judgments} \\ \textcolor{red}{($-$ analysis} | \textcolor{teal}{Difference} \\ \textcolor{teal}{experiences ($-$ contrasting} \\ \textcolor{teal}{experience}{]} "\{input\}" "\end{tabular}                                           & \textbf{73.8 \scriptnumber{0.1}}                        & 94.0 \scriptnumber{0.2}                              & 89.2 \scriptnumber{0.2}                                & 62.6 \scriptnumber{1.1}                                   & 85.2 \scriptnumber{0.1}                                    & \textbf{25.6 \scriptnumber{0.1}}                    & \textbf{59.9} \scriptnumber{0.1}                         & 33.5 \scriptnumber{0.5}                            \\ \bottomrule
\end{tabular} 
}
\caption{Text style transfer performance for various baseline and learned prompts. Manual refers to manually-written prompts, with 1 from \cite{reif2021recipe} and 2-3 written for this experiment. Fluent refers to prompts learned using our method with fluency constraint (\S\ref{subsec:analysis}). RL refers to our main prompt optimization method. The metrics are the same as in Table \ref{tab:tst-yelp-auto}. All outputs are generated using GPT-2-xl and metrics are averaged over 5 runs. Numbers in (parentheses) are standard errors of the averaged metrics.
}
\label{tab:tst-prompt-examples}
\end{table*}

\begin{table*}[h]
\setlength{\tabcolsep}{2.5pt}
\centering
{\renewcommand{\arraystretch}{1.2}
\small
\begin{tabular}{llrrrrrrrr} \toprule
ID                                                  
& {\begin{tabular}[c]{@{}l@{}}Template \\ {[}\textcolor{red}{to old} | {\textcolor{teal}{to modern}}{]}\end{tabular}}                                                                                                            & \multicolumn{1}{l}{{Content}} & \multicolumn{1}{l}{{Style}} & \multicolumn{1}{l}{{Fluency}} & \multicolumn{1}{l}{{{\bf $\bm{J}$({\scriptsize C, S, F})}}} & \multicolumn{1}{l}{\textbf{GM({\scriptsize C, S, F})}} & \multicolumn{1}{l}{{BLEU}} & \multicolumn{1}{l}{{BERTScore}} & \multicolumn{1}{l}{{PPL$\downarrow$}} \\ \midrule
\rowcolor{Gray}
\multicolumn{10}{l}{\textit{Null Prompt}}                   \\
1                                                 
& "\{input\}" "                                                                                                                                                                                           & 41.9 \scriptnumber{0.6}                                & 56.1 \scriptnumber{1.3}                             & \textbf{87.6 \scriptnumber{0.3}}                        & 17.3 \scriptnumber{0.3}                                  & 59.0 \scriptnumber{0.2}                                   & 9.3 \scriptnumber{0.2}                              & 32.7 \scriptnumber{0.3}                                 & \textbf{48.1 \scriptnumber{0.4}}                            \\ 
\vspace{0cm} \\
\rowcolor{Gray}
\multicolumn{10}{l}{\textit{Manual Prompt}}                   \\
 1                                            
& \begin{tabular}[c]{@{}l@{}}Here is some text: "\{input\}". \\ Here is a rewrite of the text, \\ which is {[}\textcolor{red}{old} |  \textcolor{teal}{modern}{]} \\ English: "\end{tabular}                                              & 61.5 \scriptnumber{0.2}                               & 51.0 \scriptnumber{1.1}                             & 80.1 \scriptnumber{0.1}                               & 22.6 \scriptnumber{0.6}                                  & {63.1 \scriptnumber{0.5}}                            & \textbf{14.6 \scriptnumber{0.1}}                             & \textbf{40.9 \scriptnumber{0.1}}                                 & {62.6 \scriptnumber{0.2}}                 \\ 
\vspace{0cm} \\
 2                                              & \begin{tabular}[c]{@{}l@{}}Change the following sentence \\ from {[}\textcolor{red}{modern} | \textcolor{teal}{old}{]} English \\ to {[}\textcolor{red}{old} | \textcolor{teal}{modern}{]} English but \\ keep its semantics. "\{input\}" "\end{tabular} & 56.0 \scriptnumber{0.9}                              & 54.1 \scriptnumber{2.3}                             & 83.3 \scriptnumber{0.4}                                & 21.4 \scriptnumber{0.8}                                  & 63.1 \scriptnumber{0.7}                                   & 13.4 \scriptnumber{0.3}                            & 39.7 \scriptnumber{0.3}                                & 61.8 \scriptnumber{0.9}                           \\ 
 \vspace{0cm} \\
 3                                              & \begin{tabular}[c]{@{}l@{}}"\{input\}". Rewrite the sentence \\ to be {[}\textcolor{red}{old} | \textcolor{teal}{new}{]} English \\ but have the same meaning. "\end{tabular}                                                               & \textbf{58.9 \scriptnumber{0.7}}                                & 53.5 \scriptnumber{2.4}                              & 83.2 \scriptnumber{0.6}                                & 22.5 \scriptnumber{1.1}                                   & 63.9 \scriptnumber{0.9}                                    & 13.9 \scriptnumber{0.3}                             & 40.7 \scriptnumber{0.2}                                  & 62.8 \scriptnumber{0.7}                            \\ 
 \vspace{0cm} \\
\rowcolor{Gray}
\multicolumn{10}{l}{\textbf{\textit{\modelname (Ours)}}}                   \\
1
& \begin{tabular}[c]{@{}l@{}}{[}\textcolor{red}{Measure$\cdot$Psal Sanskrit thereto}$^*$ | \\ \textcolor{teal}{TacomaExcellent happiness} \\ \textcolor{teal}{verbs positives{}}{]} "\{input\}" "\end{tabular}                                                     & 49.9 \scriptnumber{0.1}                                & \textbf{67.3 \scriptnumber{0.4}}                      & 85.3 \scriptnumber{0.1}                                & {26.4 \scriptnumber{0.1}}                           & 65.9 \scriptnumber{0.1}                                    & 12.6 \scriptnumber{0.1}                             & 38.0 \scriptnumber{0.1}                                  & 64.5 \scriptnumber{0.4}                            \\ 
\vspace{0cm} \\
2
& \begin{tabular}[c]{@{}l@{}}{[}\textcolor{red}{Character Psal Quran verbsð} | \\ \textcolor{teal}{ Verb Effect verb Effect verb}{]} \\ "\{input\}" "\end{tabular}                                                 & 52.2 \scriptnumber{0.0}                                & 61.7 \scriptnumber{0.4}                              & 85.0 \scriptnumber{0.2}                                & 25.4 \scriptnumber{0.1}                                   & 64.9 \scriptnumber{0.1}                                    & 13.3 \scriptnumber{0.1}                             & 39.0 \scriptnumber{0.1}                                  & 63.2 \scriptnumber{0.3}                            \\
\vspace{0cm} \\ 
3
& \begin{tabular}[c]{@{}l@{}}{[}\textcolor{red}{ search (< Psal Ethiop} \\ \textcolor{red}{differentiate} | \textcolor{teal}{ Meaning Usage} \\ \textcolor{teal}{ phr phr phr}{]} "\{input\}" "\end{tabular}                                           & 53.3 \scriptnumber{0.1}                       & 66.3 \scriptnumber{0.3}                              & 85.3 \scriptnumber{0.1}                                & \textbf{28.3 \scriptnumber{0.1}}                                   & \textbf{67.1 \scriptnumber{0.1}}                                    & {13.3} \scriptnumber{0.0}                    & {39.9} \scriptnumber{0.1}                         & 61.9 \scriptnumber{0.3}                            \\ \bottomrule
\end{tabular} 
}
\caption{Text style transfer performance for various baseline and learned prompts on Shakespeare \cite{xu-etal-2012-shakespeare}. The metrics and format are the same as Table \ref{tab:tst-prompt-examples}. 
$^*$The dot in this prompt should be the ``dagesh'' character in Hebrew, with unicode number U+05BC. Here we use \textbackslash{}cdot for easier rendering.
}
\label{tab:tst-prompt-examples-shakespeare}

\end{table*}

\end{document}

%% file: tab-NLU-main-result.tex
\begin{table*}[t]
\vspace{-5pt}
\centering
\begin{center}
{\renewcommand{\arraystretch}{1.0}
\resizebox{1\textwidth}{!}{
\setlength{\tabcolsep}{4pt}
\small
\begin{tabular}{@{}lllllllll@{}}
\toprule
                      & \textbf{SST-2}     & \textbf{Yelp P.} & \textbf{MR}    & \textbf{CR}  & \textbf{SST-5} & \textbf{Yelp} &  \textbf{AG's News}   & \textbf{Avg.} \\ \midrule
 Fine-Tuning          & 80.6 \scriptnumber{3.9}         & 88.7 \scriptnumber{4.7}       & 67.4 \scriptnumber{9.7}     & 73.3 \scriptnumber{7.5} & 40.7 \scriptnumber{3.0} & \textbf{51.0} \scriptnumber{2.2}  & \textbf{84.9} \scriptnumber{3.6}          &  69.5  \\ 
 Manual Prompt       & 82.8               & 83.0             & 80.9           & 79.6    & 34.9 & 42.1      & 76.9               &  68.6  \\
 Instructions        & 89.0               & 84.4             & 85.2           & 80.8     & 29.8 & 43.0    & 54.8               &  58.5  \\
 In-Context Demonstration    & 85.9 \scriptnumber{0.7}         & 89.6 \scriptnumber{0.4}       & 80.6 \scriptnumber{1.4}     & 85.5 \scriptnumber{1.5} & 39.3 \scriptnumber{0.9} & 49.4 \scriptnumber{0.3}  & 74.9 \scriptnumber{0.8}   &  72.2  \\
 Prompt Tuning \emph{(Soft Prompt Tuning)}        & 73.8 \scriptnumber{10.9}        & 88.6 \scriptnumber{2.1}       & 74.1 \scriptnumber{14.6}    & 75.9 \scriptnumber{11.8} & 40.2 \scriptnumber{6.5} & 49.1 \scriptnumber{3.1}   & 82.6 \scriptnumber{0.9}      &  69.2  \\
 BB Tuning \emph{(2 soft tokens)}      & 83.2 \scriptnumber{3.5}         & 86.0 \scriptnumber{1.6}       & 77.1 \scriptnumber{3.9}     & 83.2 \scriptnumber{2.5} & 39.2 \scriptnumber{2.4}  &  41.5 \scriptnumber{1.9}   & 74.0 \scriptnumber{1.9}         &  69.2  \\
 BB Tuning \emph{(5 soft tokens)}      & 84.6 \scriptnumber{4.0}         & 78.7 \scriptnumber{2.3}       & 79.8 \scriptnumber{1.5}     & 82.9 \scriptnumber{3.6} &  36.6 \scriptnumber{2.1} & 33.7 \scriptnumber{2.3}    & 73.6 \scriptnumber{3.6}         &  67.1  \\
 BB Tuning \emph{(Mixed, 50 soft tokens)}      & 89.1 \scriptnumber{0.9}         & 93.2 \scriptnumber{0.5}       & 86.6 \scriptnumber{1.3}     & 87.4 \scriptnumber{1.0} &  38.4 \scriptnumber{1.1} & 44.8 \scriptnumber{1.3}   & 83.5 \scriptnumber{0.9}         &  74.7  \\
 GrIPS \emph{(Discrete Prompt Enumeration)}                 & 87.1 \scriptnumber{1.5}         & 88.2 \scriptnumber{0.1}       & 86.1 \scriptnumber{0.3}     & 80.0 \scriptnumber{2.5} & 32.0 \scriptnumber{1.8} & 47.2 \scriptnumber{0.5}   & 65.4 \scriptnumber{9.8}      &  69.4  \\ 
 AutoPrompt            & 75.0 \scriptnumber{7.6}         & 79.8 \scriptnumber{8.3}       & 62.0 \scriptnumber{0.8}     & 57.5 \scriptnumber{5.8} & 27.8 \scriptnumber{3.3} & 29.0 \scriptnumber{5.0}   & 65.7 \scriptnumber{1.9}          &  56.7  \\

 \midrule
 RLPrompt (Ours, 2 discrete tokens)            &       90.3 \scriptnumber{1.3}             &     94.1 \scriptnumber{0.8}              &  86.5 \scriptnumber{1.2}              &     87.4 \scriptnumber{1.7}         & 40.1 \scriptnumber{1.9}    & 45.6 \scriptnumber{3.8} &     76.8 \scriptnumber{1.4}  & 74.4    \\ 
 RLPrompt (Ours, 5 discrete tokens)            &       \textbf{92.5} \scriptnumber{0.8}             &     \textbf{95.1} \scriptnumber{1.0}              &  \textbf{87.1} \scriptnumber{0.4}              &     \textbf{89.5} \scriptnumber{0.6}         & \textbf{41.4} \scriptnumber{3.2}   & 44.8 \scriptnumber{4.3} &      80.2 \scriptnumber{0.7} &   \textbf{75.8}  \\

 \bottomrule

 \end{tabular}}}
 
\caption{\small Results of few-shot text classification. The last column shows the average accuracy across all datasets in this table. Additional results can be found in Table~\ref{tab:cls-addition}.
}
\label{tab:cls-main}
\end{center}
\end{table*}

%% file: fig-training-efficiency.tex
\begin{figure}[t]
    \vspace{-6pt}
    \centering
    \includegraphics[width=0.48\textwidth]{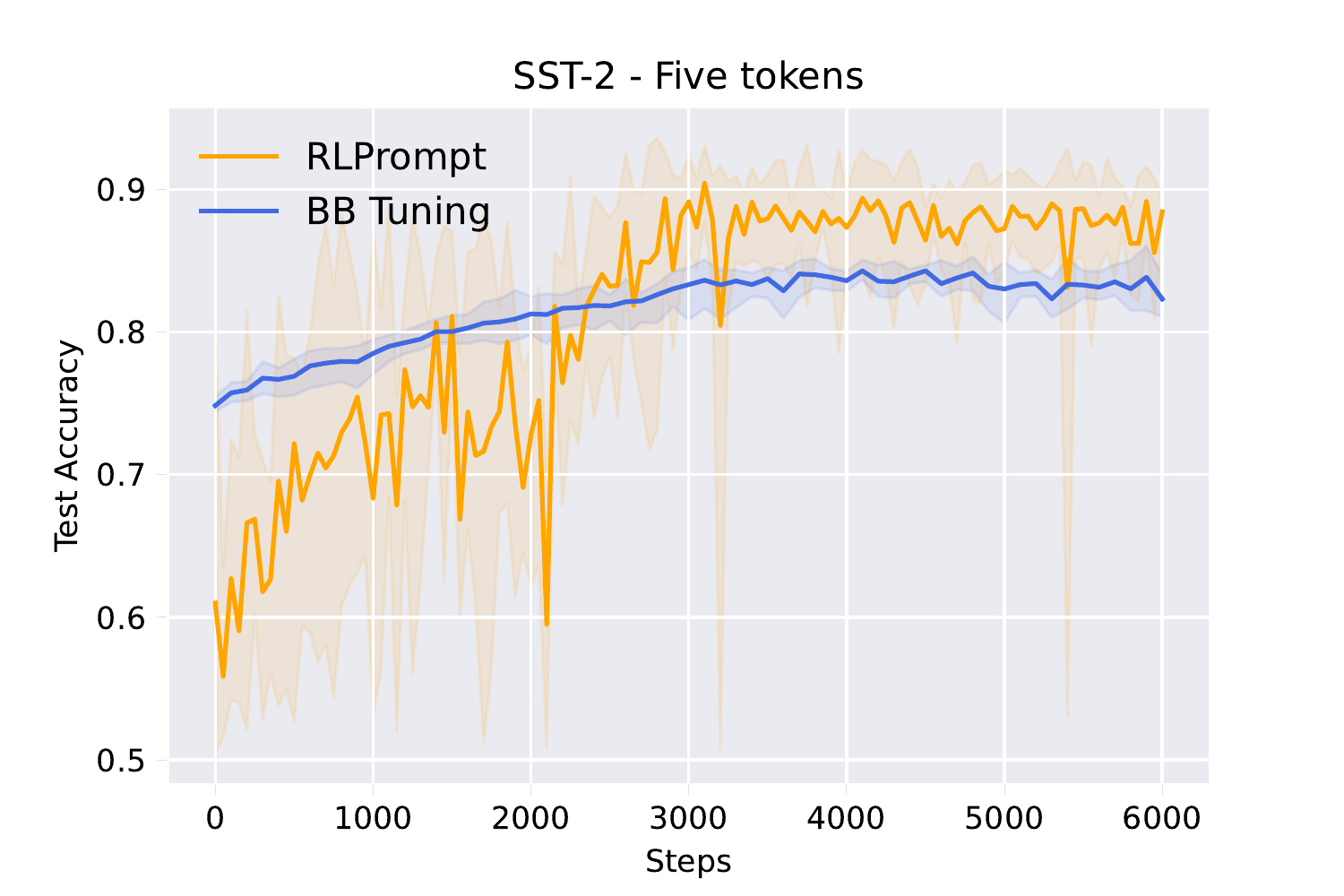}
    \vspace{-20pt}
    \caption{\small Comparison of our method ({\color{orange} orange}) and Black-Box (BB) Tuning~\citep{sun2022black} ({\color{blue} blue}) in terms of training efficiency. The solid curves are the mean and the shaded regions are the maximum and minimum test accuracy over 5 trials.}
    \label{fig:train-efficiency}
\end{figure}

%% file: tab-NLU-dataset.tex
\begin{table*}[t]
\centering
\resizebox{\textwidth}{!}{
\begin{tabular}{@{}lllllll@{}}
\toprule
\textbf{Dataset}   & \textbf{Type}                & $|C|$               & \textbf{|Train|}=\textbf{|Dev|}        & \textbf{|Test|}               & \textbf{Manual template}                                                            & \textbf{Label words}                                    \\
\toprule
SST-2     & Sentiment (Movie reviews)   & 2                    & $16 \times |C|$                   & 1.8k                 & \texttt{\textless{}S\textgreater{}} It was \texttt{[MASK]} .                       & terrible, great                                \\
\vspace{-0.2cm} \\
Yelp P.   & Sentiment (Yelp reviews)   & 2                    & $16 \times |C|$                   & 38k                  & \texttt{\textless{}S\textgreater{}} It was \texttt{[MASK]} .                       & terrible, great                                \\
\vspace{-0.2cm} \\
MR        & Sentiment (Movie reviews)   & 2                    & $16 \times |C|$                   & 2k                & \texttt{\textless{}S\textgreater{}} It was \texttt{[MASK]} .                       & terrible, great                                \\
\vspace{-0.2cm} \\
CR        & Sentiment (Product reviews)   & 2                    & $16 \times |C|$                   & 2k                   & \texttt{\textless{}S\textgreater{}} It was \texttt{[MASK]} .                       & terrible, great                                \\
\vspace{-0.2cm} \\
SST-5     & Sentiment (Movie reviews)   & 5                    & $16 \times |C|$                   & 2.2k                 & \texttt{\textless{}S\textgreater{}} It was \texttt{[MASK]} .                       & terrible, bad, okay, good, great                                \\
\vspace{-0.2cm} \\
Yelp      & Sentiment (Yelp reviews)   & 5                    & $16 \times |C|$                   & 50k                  & \texttt{\textless{}S\textgreater{}} It was \texttt{[MASK]} .                       & terrible, bad, okay, good, great                                \\
\vspace{-0.2cm} \\
Subj & Subjectivity (Movie reviews) & 2 & $16 \times |C|$                   & 2k & \texttt{\textless{}S\textgreater{}} This is \texttt{[MASK]} . & subjective, objective \\
\vspace{-0.2cm} \\
AG's News & Topic (News articles)       & 4                    & $16 \times |C|$                   & 7.6k                 & \texttt{[MASK]} News: \texttt{\textless{}S\textgreater{}}                          & World, Sports, Business, Tech                  \\
\vspace{-0.2cm} \\
TREC & Topic (Question types) & 6 & $16 \times |C|$                   & 0.5k & \texttt{[MASK]}: \texttt{\textless{}S\textgreater{}} & 
\begin{tabular}[c]{@{}l@{}}Description, Entity, Expression, Human, \\ Location, Number \end{tabular} \\
\vspace{-0.2cm} \\
DBPedia   & Topic (Wikipedia ontologies)       & 14                   & $16 \times |C|$                   & 70k                  & \texttt{[{}Category: [MASK]]} \texttt{\textless{}S\textgreater{}}                & 
\begin{tabular}[c]{@{}l@{}}Company, Education, Artist, Sports, Office, \\ Transportation, Building, Natural, Village, \\ Animal, Plant, Album, Film, Written\end{tabular} \\
\vspace{-0.2cm} \\
Yahoo     & Topic (Question types)       & 10                   & $16 \times |C|$                   & 60k                  & Topic \texttt{[MASK]}: \texttt{\textless{}S\textgreater{}}  &
\begin{tabular}[c]{@{}l@{}}culture, science, health, education, computer, \\ sports, business, music, family, politics\end{tabular} \\


\bottomrule
\end{tabular}}
\caption{Main datasets evaluated in this work. $|C|$: $\#$ of classes for classification tasks. \texttt{\textless{}S\textgreater{}}: input sentence. All our label words have a prepended special character Ġ to represent a space before a word. Note that we follow the true few-shot learning setting \cite{perez2021trueFS} by taking the same number of validation and training, which is consistent with previous prompting works. 
}
\label{tab:nlu-dataset}
\end{table*}

%% file: tab-NLU-additional-result.tex
\begin{table*}[h]
\centering
\begin{center}
{\renewcommand{\arraystretch}{1.0}
\small
\vspace{10pt}
\begin{tabular}{@{}llllll@{}}
\toprule
& Subj                               & TREC                & Yahoo               & DBPedia  & Avg.           \\
\midrule
Fine-Tuning                 & \textbf{89.0} \scriptnumber{3.5}                & \textbf{83.9} \scriptnumber{5.5} & \textbf{65.6} \scriptnumber{2.4} & \textbf{97.7} \scriptnumber{0.8} & \textbf{84.1} \\
Manual Prompt               & 51.5                               & 31.8                & 18.1                & 59.2  & 40.2              \\
Instructions                & 50.4                               & 26.2                & 21.4                & 15.9  & 28.5              \\
In-Context Demonstration             & 51.9 \scriptnumber{1.3}                         & 29.2 \scriptnumber{2.0}          & 36.7 \scriptnumber{2.1}          & 76.6 \scriptnumber{0.4}  & 48.6        \\
Prompt Tuning \textit{(Soft Prompt Tuning)}               & 73.0 \scriptnumber{7.3}                         & 49.6 \scriptnumber{6.1}          & \underline{59.7} \scriptnumber{1.3}          & 84.2 \scriptnumber{5.3}   & 66.6       \\
BB Tuning \textit{(2 soft tokens)}          & 75.7 \scriptnumber{3.4}                         & 40.4 \scriptnumber{2.5}          & 41.7 \scriptnumber{1.4}          & 60.9 \scriptnumber{6.0}  & 54.7        \\
BB Tuning \textit{(5 soft tokens)}         & 75.8 \scriptnumber{4.4}                         & 39.8 \scriptnumber{4.6}          & 38.2 \scriptnumber{1.8}          & 62.7 \scriptnumber{4.1}  & 54.1        \\
BB Tuning \textit{(Mixed, 50 soft tokens)} & 71.8 \scriptnumber{5.1}                         & 46.4 \scriptnumber{8.2}          & 50.0 \scriptnumber{0.9}          & \underline{90.2} \scriptnumber{0.8}  & 64.6        \\
GrIPS \textit{(Discrete Prompt Enumeration)}                      & 74.8 \scriptnumber{1.1}                         & 9.5 \scriptnumber{0.2}           & 22.5 \scriptnumber{0.4}          & 22.1 \scriptnumber{2.9} & 32.2         \\
AutoPrompt                  & 78.9 \scriptnumber{4.5}                         & 38.8 \scriptnumber{4.3}          & 35.5 \scriptnumber{2.0}          & 63.1 \scriptnumber{2.0}  & 54.1        \\
\midrule
RLPrompt (2 discrete tokens)                & \underline{81.9} \scriptnumber{1.2}              & \underline{60.5} \scriptnumber{3.3}          & 48.6 \scriptnumber{0.6}          & 76.0 \scriptnumber{0.6} & 66.8         \\
RLPrompt (5 discrete tokens)                & 81.2 \scriptnumber{1.7}                         & 57.6 \scriptnumber{4.6}          & 48.6 \scriptnumber{1.0}          & 84.6 \scriptnumber{1.9} & \underline{68.0}         \\
\bottomrule
\end{tabular}}
\caption{Additional results of few-shot text classification. The best result on each dataset is {\bf bolded} and the second best result \underline{underscored}. The remaining format follows Table~\ref{tab:cls-main}.}
\label{tab:cls-addition}
\end{center}
\end{table*}

%% file: tab-NLU-instruction.tex
\begin{table*}[t]
\centering
\resizebox{\textwidth}{!}{
\begin{tabular}{rp{16cm}}
\toprule
Dataset & SST-2 \\ 
Instruction & In this task, you are given sentences from movie reviews. The task is to classify a sentence as "great" if the sentiment of the sentence is positive or as "terrible" if the sentiment of the sentence is negative. \\
\modelname 2 token template & \texttt{\textless{}S\textgreater{}} VERY Absolutely \texttt{[MASK]} .\\
\modelname 5 token template & \texttt{\textless{}S\textgreater{}} AgentMediaGradeOfficials Grade
\texttt{[MASK]} .\\
\midrule
Dataset & Yelp P. \\ 
Instruction & In this task, you are given Yelp reviews. The task is to classify a review as "great" if the overall sentiment of the review is positive or as "terrible" if the overall sentiment of the review is negative. \\
\modelname 2 token template & \texttt{\textless{}S\textgreater{}}  Rating Absolutely \texttt{[MASK]} .\\
\modelname 5 token template & \texttt{\textless{}S\textgreater{}}  ProductGradeTimeoutAbsolutely Absolutely \texttt{[MASK]} .\\
\midrule
Dataset & MR \\
Instruction & In this task, you are given sentences from movie reviews. The task is to classify a sentence as "great" if the sentiment of the sentence is positive or as "terrible" if the sentiment of the sentence is negative \\
\modelname 2 token template & \texttt{\textless{}S\textgreater{}}  downright absolutely \texttt{[MASK]} .\\
\modelname 5 token template & \texttt{\textless{}S\textgreater{}} ouslyicals downright certainly consistently \texttt{[MASK]} .\\
\midrule
Dataset & CR \\
Instruction & In this task, you are given sentences from customer reviews. The task is to classify a sentence as "great" if the sentiment of the sentence is positive or as "terrible" if the sentiment of the sentence is negative. \\
\modelname 2 token template & \texttt{\textless{}S\textgreater{}}  ITNESSALLY \texttt{[MASK]} .\\
\modelname 5 token template & \texttt{\textless{}S\textgreater{}}  absoluteliterally absolute downright downright \texttt{[MASK]} .\\
\midrule
Dataset & SST-5 \\ 
Instruction & In this task, you are given sentences from movie reviews. Based on the given review, classify it to one of the five classes: (1) terrible, (2) bad, (3) okay, (4) good, and (5) great. \\
\modelname 2 token template & \texttt{\textless{}S\textgreater{}} Movie entirely \texttt{[MASK]} .\\
\modelname 5 token template & \texttt{\textless{}S\textgreater{}} iciticititableually immediately \texttt{[MASK]} .\\
\midrule
Dataset & Yelp \\ 
Instruction & In this task, you are given Yelp reviews. Based on the given review, classify it to one of the five classes: (1) terrible, (2) bad, (3) okay, (4) good, and (5) great. \\
\modelname 2 token template & \texttt{\textless{}S\textgreater{}} =-=- Totally \texttt{[MASK]} .\\
\modelname 5 token template & \texttt{\textless{}S\textgreater{}} imalimalimalivable Totally \texttt{[MASK]} .\\
\midrule
Dataset & AG's News \\
Instruction & In this task, you are given a news article. Your task is to classify the article to one out of the four topics "World", "Sports", "Business", "Tech" if the article"s main topic is relevant to the world, sports, business, and technology, correspondingly. If you are not sure about the topic, choose the closest option. \\
\modelname 2 token template & \texttt{[MASK]} Reviewer Information \texttt{\textless{}S\textgreater{}} .\\
\modelname 5 token template & \texttt{[MASK]} StaffAreaFocusHardware Advisory \texttt{\textless{}S\textgreater{}} .\\
\midrule
Dataset & Subj \\
Instruction &  In this task, you are given sentences from reviews. The task is to classify a sentence as "subjective" if the opinion of the sentence is subjective or as "objective" if the opinion of the sentence is objective. \\
\modelname 2 token template & \texttt{\textless{}S\textgreater{}} Friends pleasantly \texttt{[MASK]} .\\
\modelname 5 token template & \texttt{\textless{}S\textgreater{}} BufferActionDialogDialog downright \texttt{[MASK]} .\\
\midrule
Dataset & TREC \\
Instruction &  You are given a question. You need to detect which category better describes the question. Answer with "Description", "Entity", "Expression", "Human", "Location", and "Number". \\
\modelname 2 token template & \texttt{\textless{}S\textgreater{}} DeveloperTermin \texttt{[MASK]} .\\
\modelname 5 token template & \texttt{\textless{}S\textgreater{}} BufferHttpRuntimeRunnerostics \texttt{[MASK]} .\\
\midrule
Dataset & Yahoo \\
Instruction &  You are given a passage. Using the information present in the passage, you need to classify it into one of the 10 topics: 0 - Culture, 1 - Science, 2 - Health, 3 - Education, 4 - Computers, 5 - Sports, 6 - Business, 7 - Music, 8 - Family, 9 - Politics. \\
\modelname 2 token template & \texttt{\textless{}S\textgreater{}} Source Ireland \texttt{[MASK]} .\\
\modelname 5 token template & \texttt{\textless{}S\textgreater{}} AlertSource mentioning Besidesadays \texttt{[MASK]} .\\
\midrule
Dataset & DBPedia \\
Instruction &  You are given a passage. Using the information present in the passage, you need to classify it into one of the 10 topics: 0 - Culture, 1 - Science, 2 - Health, 3 - Education, 4 - Computers, 5 - Sports, 6 - Business, 7 - Music, 8 - Family, 9 - Politics.  \\
\modelname 2 token template &  typeSection \texttt{[MASK]} : \texttt{\textless{}S\textgreater{}} .\\
\modelname 5 token template &  CommonExamplesSenate Similar comparable \texttt{[MASK]} : \texttt{\textless{}S\textgreater{}} .\\
\bottomrule
\end{tabular}}
\caption{Manual instructions (following \textit{natural instructions} \cite{mishra2021NI}) we tested with in our baseline implementation and some template cases we learned by \modelname for specific datasets.}

\label{tab:nlu-instruction}
\end{table*}